%% file: neurips_2025.tex
\pdfoutput=1
\documentclass{article}

   \PassOptionsToPackage{numbers, compress}{natbib}
\usepackage[preprint]{neurips_2025}

\usepackage{graphicx}
\usepackage{caption}
\usepackage{subfig}
\usepackage[table,xcdraw]{xcolor}
\usepackage{makecell}
\usepackage{booktabs}
\usepackage{array}

\usepackage{graphicx}
\usepackage{float}
\usepackage{lipsum}
\usepackage{multirow}
\usepackage{pifont}
\usepackage{amsmath}

\usepackage[utf8]{inputenc} % allow utf-8 input
\usepackage[T1]{fontenc}    % use 8-bit T1 fonts
\usepackage{hyperref}       % hyperlinks
\usepackage{url}            % simple URL typesetting
\usepackage{booktabs}       % professional-quality tables
\usepackage{amsfonts}       % blackboard math symbols
\usepackage{nicefrac}       % compact symbols for 1/2, etc.
\usepackage{microtype}      % microtypography

\title{Text-Aware Image Restoration with Diffusion Models}

\author{
    Jaewon Min$^{1\ast}$ \qquad Jin Hyeon Kim$^{2\ast}$ \qquad Paul Hyunbin Cho$^{1}$ \vspace{0.3em}\\
    \qquad \textbf{Jaeeun Lee}$^{3}$ \qquad \textbf{Jihye Park}$^{4}$ \qquad \textbf{Minkyu Park}$^{4}$ \vspace{0.3em} \\ 
    \textbf{Sangpil Kim}$^{2\dagger}$ \qquad \textbf{Hyunhee Park}$^{4\dagger}$ \qquad \textbf{Seungryong Kim}$^{1\dagger} $ \vspace{0.3em} \\
    $^{1}$KAIST AI \qquad $^{2}$Korea University \vspace{0.3em} \qquad $^{3}$Yonsei University \qquad $^{4}$Samsung Electronics
}

\newcommand{\ourdataset}{SA-Text}
\newcommand{\ourmodel}{TeReDiff}
\newcommand{\cmark}{\textcolor{green!70!black}{\ding{51}}}
\newcommand{\xmark}{\textcolor{red}{\ding{55}}}

\newcommand{\tablestyle}[2]{%
  \setlength{\tabcolsep}{#1}\renewcommand{\arraystretch}{#2}
}

\usepackage{wrapfig}

\begin{document}

\begingroup
\renewcommand{\thefootnote}{}
\footnotetext{$^\ast$: Equal contribution}
\footnotetext{$^\dagger$: Corresponding authors}
\endgroup

\maketitle

\input{figs/teaser}

\input{sections/0_abstract}
\input{sections/1_intro}
\input{sections/2_related}

\input{sections/3_ourdataset}
\input{sections/4_method}

\input{sections/5_experiment}
\input{sections/6_conclusion}

%\input{sections/checklist}

%%%%%%%%%%%%%%%%%%%%%%%%%%%%%%%%%%%%%%%%%%%%%%%%%%%%%%%%%%%%
\newpage
\appendix
\input{sections/supple}

%%%%%%%%%%%%%%%%%%%%%%%%%%%%%%%%%%%%%%%%%%%%%%%%%%%%%%%%%%%%

\newpage
\clearpage
\bibliography{neurips_2025}

\end{document}

%% file: figs/teaser.tex
\begin{figure}[h]
    \centering
    \includegraphics[width=\linewidth]{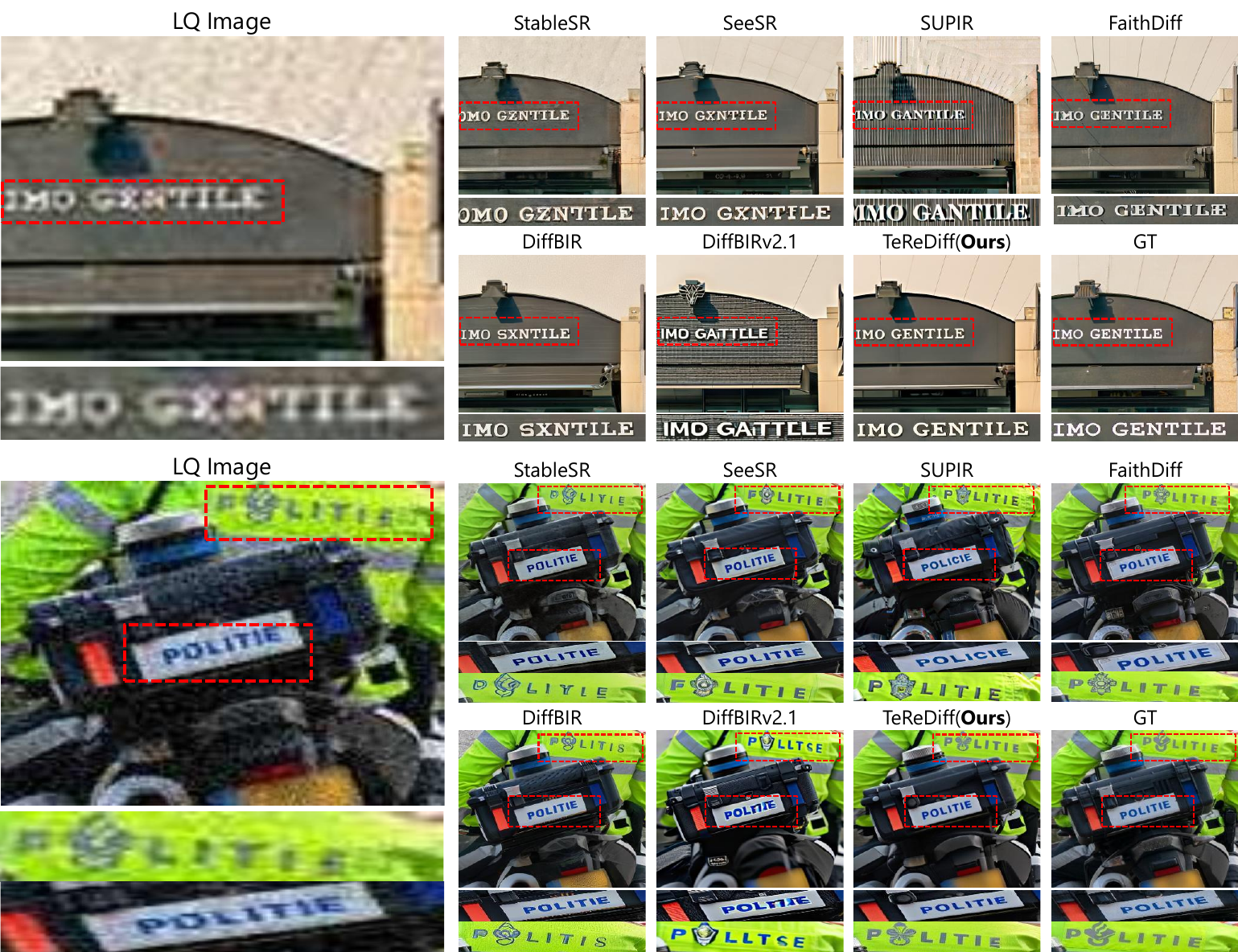}
    \vspace{-10pt}
    \caption{
    \textbf{Text-Aware Image Restoration (TAIR).} Given a low-quality (LQ) image containing degraded text, our method faithfully restores the original textual content with high legibility and fidelity, whereas previous diffusion-based models~\cite{blattmann2023stable, yu2024scaling, lin2024diffbir, chen2024faithdiff} often fail to recover the text regions. 
    %Additional qualitative results are provided in the supplementary material.
    }
    \label{fig:teaser}
\end{figure}

%% file: sections/0_abstract.tex
\begin{abstract}
Image restoration aims to recover degraded images. However, existing diffusion-based restoration methods, despite great success in natural image restoration, often struggle to faithfully reconstruct textual regions in degraded images. Those methods frequently generate plausible but incorrect text-like patterns, a phenomenon we refer to as text–image hallucination. In this paper, we introduce \textbf{Text-Aware Image Restoration (TAIR)}, a novel restoration task that requires the simultaneous recovery of visual contents and textual fidelity. To tackle this task, we present \textbf{\ourdataset}, a large-scale benchmark of 100K high-quality scene images densely annotated with diverse and complex text instances. Furthermore, we propose a multi-task diffusion framework, called \textbf{\ourmodel}, that integrates internal features from diffusion models into a text-spotting module, enabling both components to benefit from joint training. This allows for the extraction of rich text representations, which are utilized as prompts in subsequent denoising steps. Extensive experiments demonstrate that our approach consistently outperforms state-of-the-art restoration methods, achieving significant gains in text recognition accuracy. See our project page: \href{https://cvlab-kaist.github.io/TAIR/}{https://cvlab-kaist.github.io/TAIR/}
\end{abstract}

%% file: sections/1_intro.tex
\section{Introduction}
\label{intro}

Image restoration is a fundamental task in computer vision, aiming to recover high-quality images from degraded observations. This task is crucial for applications ranging from photography enhancement to medical and autonomous‑vision systems. Recent advances in generative modeling—particularly diffusion models~\cite{ho2020denoising, rombach2022high}—have demonstrated remarkable capabilities in image restoration by leveraging powerful generative priors, achieving superior perceptual quality across various degradation scenarios~\cite{yue2023resshift, wang2024exploiting, lin2024diffbir, wu2024seesr, yu2024scaling, chen2024faithdiff}. 

However, previous models still struggle to recover text regions, as shown in Fig.~\ref{fig:teaser}. Since these models rely on powerful generative priors of diffusion models~\cite{lin2024diffbir, wu2024seesr, yu2024scaling, chen2024faithdiff}, they often synthesize \textit{plausible} textures instead of reconstructing the precise characters, leading to a \textit{text-image hallucination}. Yet, textual content provides semantic cues that are essential for scenarios such as document digitization~\cite{ma2018docunet, baek2019wrong, li2023foreground, feng2023deep, feng2025docscanner}, street sign understanding~\cite{zhu2016traffic, tabernik2019deep, ertler2020mapillary}, or AR navigation~\cite{li2020textslam}, where even slight distortions can compound into significant information loss. Most image restoration studies have focused on the overall perceptual quality and have not explicitly addressed text readability.

On the other hand, image super‑resolution methods for text restoration have attempted to improve the perceptual quality and legibility of \textit{cropped} text regions~\cite{dong2015boosting,wang2019textsr,xu2017learning,ledig2017photo}. However, their patch‑level focus introduces fundamental limitations. The global context is discarded by focusing solely on cropped text regions, thereby ignoring information crucial for overall visual coherence. Additionally, most methods train the model from scratch without leveraging large-scale generative priors, restricting their ability to generate high‑quality images.

To address these limitations, we propose a new task: \textbf{Text-Aware Image Restoration (TAIR)}. In contrast to conventional image restoration approaches~\cite{wang2021real, liang2021swinir, yue2023resshift, wang2024exploiting, lin2024diffbir, wu2024seesr, yu2024scaling, chen2024faithdiff}, TAIR necessitates the integration of textual semantics within the restoration process by operating on full natural images that contain text of varying sizes and spatial contexts—unlike prior methods~\cite{dong2015boosting,wang2019textsr,xu2017learning,ledig2017photo} that focused solely on cropped text regions. However, a primary challenge lies in the absence of suitable datasets. Existing image restoration benchmarks~\cite{agustsson2017ntire, cai2019toward, li2023lsdir, wei2020component} are not designed to address text reconstruction, making it difficult to train models that align visual restoration with text readability. While some datasets~\cite{singh2021textocr, veit2016coco,Gupta16,  karatzas2015icdar, yuliang2017detecting, ch2017total} provide image-text pairs for text-spotting, they are suboptimal for our purpose due to their limitations in scale and quality. These datasets, typically created through synthetic generation~\cite{Gupta16} or manual annotations~\cite{singh2021textocr,veit2016coco, yuliang2017detecting, ch2017total}, often suffer from low resolution.

To overcome this, we introduce a large-scale dataset, \textbf{\ourdataset}, specifically curated for TAIR. Our dataset curation pipeline begins with automatic text detection to identify candidate images containing text regions. These candidates are then validated using vision-language models~\cite{Qwen2.5-VL, lu2024ovis} (VLMs), which align semantic labels with corresponding textual regions. After filtering out low-quality images using VLMs, we obtain high-quality crops of scene images paired with accurate text annotations. Based on SA-1B~\cite{kirillov2023segment}, \ourdataset{} comprises 100K images densely annotated with rich textual content. The text in this dataset encompasses a diverse range of font styles, sizes, orientations, and complex visual contexts, offering a robust benchmark for evaluating TAIR. To the best of our knowledge, this is the first benchmark to jointly evaluate perceptual restoration quality and text fidelity.

The primary objective of TAIR is to restore full scene images while faithfully preserving the original textual content. To achieve this, we propose a new model for TAIR, named \textbf{\ourmodel} (\textbf{Te}xt \textbf{Re}storation \textbf{Diff}usion model) that combines a diffusion-based image restoration model with a text-spotting module. Inspired by recent studies demonstrating the effectiveness of diffusion features in vision downstream tasks~\cite{xu2023open, ke2024repurposing, tang2023emergent}, we directly use diffusion U‑Net features as input to the text‑spotting module. This multi-task diffusion framework improves the text spotting performance by leveraging semantically rich and text-aware representations, while also enhancing restoration performance through shared features. Furthermore, at inference time, the output of the text spotting module can be leveraged to generate input prompts for subsequent denoising steps, thereby jointly enhancing visual quality and text readability.

In summary, our key contributions are as follows:
\begin{itemize}
    \item We introduce Text‑Aware Image Restoration (TAIR), the first restoration task that explicitly requires simultaneous recovery of \emph{scene appearance} and \emph{textual fidelity}.

    \item We release \ourdataset, a dataset of 100K high-quality images densely annotated with diverse and VLM-verified text, enabling rigorous evaluation and further research on text‑conditioned restoration.
    
    \item We propose \ourmodel{}, a model trained within a multi‑task diffusion framework where U‑Net features are forwarded to a text‑spotting model during training, while the spotted text is provided as a prompt at inference time, yielding mutual gains in perceptual quality and character legibility.

\end{itemize}

%% file: sections/2_related.tex
\input{figs/data_pipeline}
\section{Related Work}

\paragraph{Diffusion-based image restoration.}
Advancements in diffusion models for high-quality image~\cite{ramesh2021zero, rombach2022high, podell2023sdxl} and video generation~\cite{yang2024cogvideox, blattmann2023stable, zheng2024open} have led to their application in various tasks, including image restoration (IR)\cite{lin2024diffbir, mei2024codi, sahak2023denoising, saharia2022image}. In contrast to previous GAN-based IR approaches\cite{wang2021real, wang2018esrgan, zhang2021blind, che2016mode, mao2019mode}, which suffer from unstable training and mode collapse, diffusion-based IR exhibits stable training, enhanced robustness, and improved generalization due to its iterative denoising nature. SR3~\cite{saharia2022image} was the first to introduce diffusion models for IR tasks, achieving state-of-the-art performance on both facial and natural image datasets. More recent works~\cite{sahak2023denoising, wang2024exploiting} have further improved DMs-based IR methods by addressing challenges related to image degradation and fidelity.

\paragraph{Text spotting.}
Scene text spotting refers to the joint task of detecting and recognizing text within natural images. Early approaches typically decomposed the problem into two distinct stages: text detection, which localizes text regions using region-based or segmentation-based techniques~\cite{liao2017textboxes, liu2017deep, ma2018arbitrary, wang2019shape, liao2020real}, and text recognition, which interprets the localized content using sequence modeling~\cite{graves2006connectionist, shi2016end, shi2018aster}. To address the challenges posed by irregularly shaped text in real-world scenes, recent detection methods have proposed polygon~\cite{ye2023dptext} or Bezier curve representations~\cite{liu2020abcnet}, substantially improving localization accuracy. Furthermore, transformer-based architectures inspired by object detection models such as DETR~\cite{carion2020end} have demonstrated strong performance in text detection~\cite{ye2023dptext, tang2022few, zhang2022text}. Concurrently, recognition has been formulated as an image-to-text translation task, with advances in visual feature extraction and cost computation~\cite{du2022svtr, lee2020recognizing, li2019show,hong2021deep,hong2022cost,hong2022neural,hong2024pf3plat,hong2024unifying,cho2021cats,cho2022cats++,cho2024cat} and language modeling~\cite{yu2020towards, fang2021read} contributing to significant improvements in recognition accuracy.

%% file: figs/data_pipeline.tex
\begin{figure*}[t]
\centerline{\includegraphics[width=1.0\textwidth]{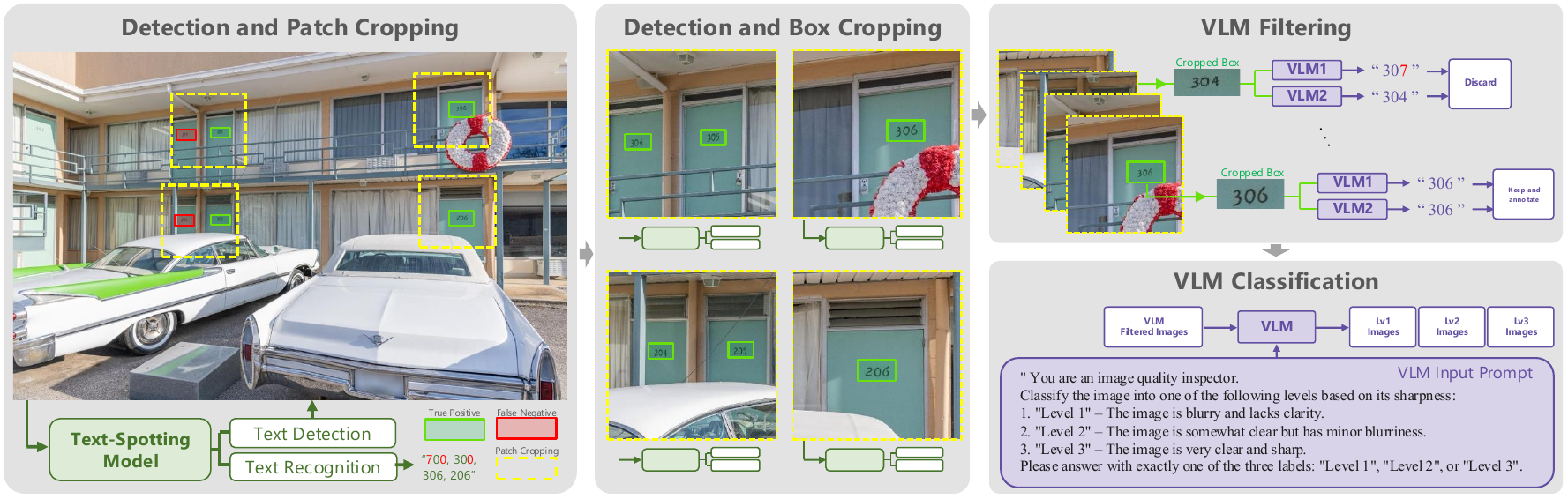}}
%\vspace{-7pt}
\caption{\textbf{\ourdataset{} curation pipeline.} 
First, a text-spotting model such as DG-Bridge Spotter~\cite{huang2024bridging} is applied to the entire image to detect text regions. Since detection at the full-image scale may fail to capture certain text instances, we further extract image patches corresponding to the detected regions and reapply the same model to each patch in order to detect potential false negatives. Next, two vision-language models (VLMs)~\cite{Qwen2.5-VL, lu2024ovis} transcribe the text within each bounding box, and only patches with consistent predictions from both models are retained and annotated. Finally, a single VLM~\cite{Qwen2.5-VL} classifies each patch into one of three categories based on image sharpness and blurriness.}
% First, a text-spotting model such as Bridge~\cite{huang2024bridging} detects texts in a given image and extracts corresponding patches. Second, using the same text-spotting model, detection is reapplied to these patches to identify false negatives from the initial stage. Third, two vision-language models (VLMs)~\cite{Qwen2.5-VL, lu2024ovis} transcribe the text within each bounding box, and only images with consistent predictions from both models are retained. Finally, a single VLM~\cite{Qwen2.5-VL} categorizes each image into three levels based on image sharpness and blurriness.}
\label{fig:data_pipeline}\vspace{-10pt}
\end{figure*}

%% file: sections/3_ourdataset.tex
\section{\ourdataset{} Dataset}
In this section, we present our \ourdataset{} curation pipeline, consisting of a detection stage and a recognition stage, as detailed in Sec.\ref{sec:dataset_pipeline}. Next, we analyze \ourdataset{} to illustrate its suitability for TAIR in Sec.\ref{sec:dataset_analysis}.
\subsection{Data Curation Pipeline}
\label{sec:dataset_pipeline}

TAIR requires datasets where textual instances in images are explicitly paired with their corresponding bounding boxes. Existing datasets like TextOCR~\cite{singh2021textocr} provide such densely annotated images for text spotting tasks. However, their low resolution makes them unsuitable for training restoration models. TextZoom~\cite{wang2020scene} provides text annotations for the high-resolution images in RealSR~\cite{cai2019toward} and SR-RAW~\cite{zhang2019zoom}, which could be leveraged for TAIR. Yet, the limited scale of TextZoom~\cite{wang2020scene} constrains the capacity for learning the fine‑grained image-text alignment, which is essential for TAIR. Furthermore, the manual construction processes of these datasets pose significant challenges for scaling to larger and more diverse datasets. 

Beyond these datasets, conventional image restoration datasets~\cite{li2023lsdir, karatzas2015icdar, lim2017enhanced}, although rich in high‑quality images, neither guarantee the presence of textual content nor provide any accompanying text annotations. In contrast, scene text spotting datasets include dense text annotations, but their images are predominantly low‑quality and thus inadequate for image restoration training. Therefore, a suitable dataset for TAIR must satisfy three critical conditions: (i) high‑quality images, (ii) dense and accurate text-location annotations, and (iii) scalability in both volume and diversity. Existing datasets do not fully meet these conditions. To bridge this gap, we propose a scalable dataset curation pipeline that produces high‑resolution images with dense text annotations, specifically tailored for TAIR. To ensure a high‑quality dataset suitable for image restoration, we sourced images from SA‑1B~\cite{kirillov2023segment}, a well‑known corpus of 11M high‑resolution images.

\input{figs/pipeline_results}

\paragraph{Text detection and region cropping.} 
The first stage of our pipeline involves running a dedicated text detection model over entire high-resolution images to identify regions containing text instances. We observe that when images are high-resolution, the detection model often misses relatively small text instances as shown in Fig.~\ref{fig:data_pipeline}. Nonetheless, based on these initial detections, we extract a 512$\times$512 crop that fully encloses at least one complete text instance and ensures that each instance appears in only one crop. As smaller text instances missed in the first pass often remain within these crops, we address this by re‑running the detection model on every crop. The reduced field of view substantially boosts recall and introduces only a minimal number of additional false positives, which are subsequently removed in our VLM‑based filtering stage. These refined detection results are passed to the recognition stage. Examples of this process are illustrated in Fig.~\ref{fig:pipeline_results}.

\paragraph{VLM-based text recognition and filtering.}
Recent vision–language models (VLMs)~\cite{Qwen2.5-VL, lu2024ovis} have exhibited strong text-recognition performance on OCRBench~\cite{liu2023hidden}, validating their effectiveness as recognition backbones. Accordingly, we adopt VLMs for our recognition stage. 
To leverage VLM accuracy without suffering from their localization shortcomings~\cite{ranasinghe2024learning}, we first isolate every text instance within a crop using bounding boxes derived from the polygons returned by the detector. Each isolated patch is submitted to two distinct VLMs to mitigate individual recognition errors and resolve ambiguities in challenging cases. We retain an instance only when both VLMs return identical transcriptions, thereby filtering out misreadings, hard‑to‑read texts, and the false positives introduced during detection. Specifically, we utilize Qwen2.5‑VL~\cite{Qwen2.5-VL} and OVIS2~\cite{lu2024ovis} in this dual-model verification process. We then apply a final filtering stage using one of the VLMs ( Qwen2.5‑VL~\cite{Qwen2.5-VL}) again to eliminate blurry or out‑of‑focus crops. SA-1B~\cite{kirillov2023segment}, despite its overall high quality, contains images with intentional blurring of human faces, license plates, and privacy-sensitive regions. Therefore, global variance‑based blur metrics, such as Laplacian filtering,  often fail to detect such localized, intentional blur. Moreover, crops may include background text that is inherently out of focus.
By employing a VLM, we successfully identify and filter both the crops that have intentionally blurred regions or naturally out-of-focus text, thereby improving dataset quality for TAIR. The result images of the overall pipeline are shown in Fig.~\ref{fig:satext_ex}. 
\input{figs/satext_examples}

\subsection{Dataset Analysis}
\label{sec:dataset_analysis}

\vspace{5pt}

\input{tables/DataQuality_table}
Leveraging our dataset curation pipeline, which is specifically designed to improve text annotation accuracy, we construct \ourdataset{}-100K from SA-1B~\cite{kirillov2023segment}. Note that the entire pipeline is fully automatic, making it readily scalable for curating even larger datasets. We compare \ourdataset{} with datasets for text spotting~\cite{veit2016coco, karatzas2015icdar, yuliang2017detecting, ch2017total, singh2021textocr} and image restoration~\cite{li2023lsdir, karatzas2015icdar, lim2017enhanced}. As shown in Tab.~\ref{tab:compre_dataset}, \ourdataset{} is the only one that provides both high-quality images and explicit text annotations, while also containing the largest number of images among all compared datasets. More detailed analyses of our dataset can be found in the supplementary materials.

%% file: figs/pipeline_results.tex
\begin{figure}[t]
    \centering
    \includegraphics[width=\linewidth]{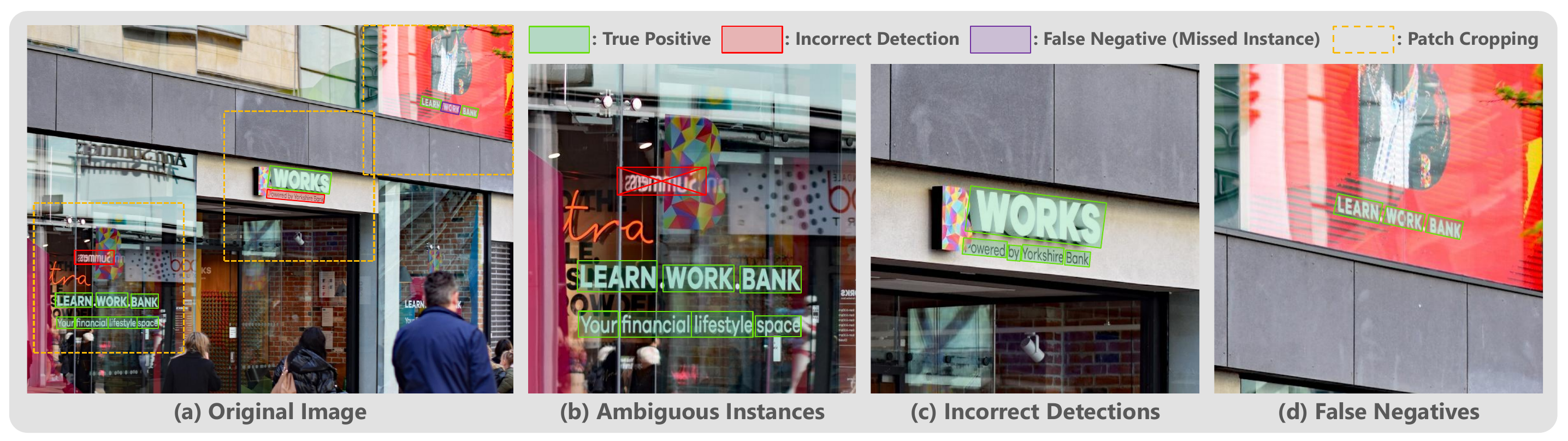}
    \vspace{-10pt}
    \caption{
    \textbf{Illustration of our dataset curation pipeline's effectiveness.} (a) Original high-resolution image with multiple text instances. (b) Ambiguous text instances are removed during the Vision–Language Model (VLM) filtering stage when the two VLMs produce differing recognition outputs. (c) Incorrect detections from the full image are corrected by re-running the detection model on smaller crops; here, the phrase "Powered by Yorkshire Bank" is successfully split into individual words. (d) False negatives (missed instances) from the initial detection on the entire image are effectively captured during the second detection pass on the smaller crops; the previously missed instance "WORK" is now correctly detected.}
    \label{fig:pipeline_results}
\end{figure}

%% file: figs/satext_examples.tex
\begin{figure}[t]
    \centering
    \includegraphics[width=\linewidth]{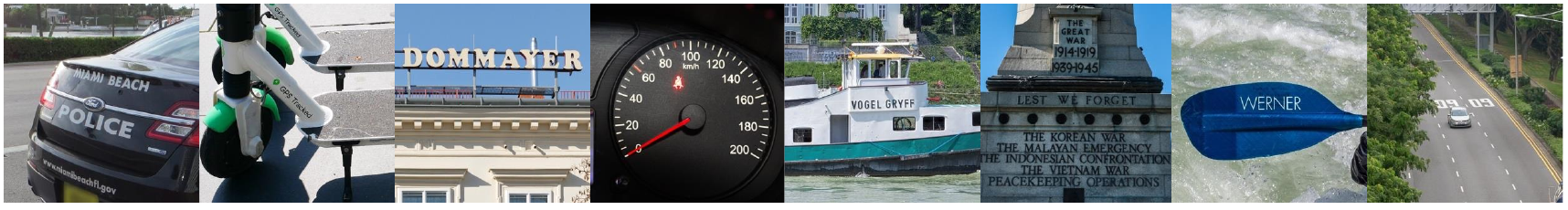}
    \vspace{-15pt}
    \caption{
    \textbf{Example images from our \ourdataset.}  
Our dataset comprises high-quality, diverse images featuring text in varied sizes, styles, and layouts—including curved, rotated, and complex forms—providing a robust foundation for the proposed TAIR task.}
    \label{fig:satext_ex}
    \vspace{-5pt}
\end{figure}

%% file: tables/DataQuality_table.tex
\begin{wraptable}{r}{0.53\textwidth} 
    \footnotesize
    \setlength{\tabcolsep}{5pt}
    \vspace{-5pt}
    \caption{\textbf{Comparison with other datasets.}}
    \resizebox{0.52\textwidth}{!}{
        \begin{tabular}{l|l|ccc}
        \toprule
        \textbf{Task} & \textbf{Dataset} & \textbf{HQ} & \textbf{Text} & \textbf{\# of Img} \\
        \midrule
        \multirow{5}{*}{\shortstack[l]{Text \\ Spotting}}
         & COCOText & \xmark & \cmark & 13,880\\
         & ICDAR2015 & \xmark & \cmark & 1,000\\
         & CTW1500   & \xmark & \cmark & 1,000 \\
         & TotalText & \xmark & \cmark & 1,255\\
         & TextOCR   & \xmark & \cmark & 21,778\\
        \cmidrule{1-5}
        \multirow{3}{*}{\shortstack[l]{Image \\ Restoration}}
         & LSDIR   & \cmark & \xmark & 84,991\\
         & DIV2K    & \cmark & \xmark & 800 \\
         & Flickr2K & \cmark & \xmark & 2,650 \\
        \cmidrule{1-5}
        \textbf{TAIR} & \textbf{\ourdataset{} (Ours)} & \cmark & \cmark & \textbf{105,330} \\
        \bottomrule
    \end{tabular}}
    \label{tab:compre_dataset}
\vspace{-10pt}
\end{wraptable}

%% file: sections/4_method.tex
\input{figs/architecture}
\input{tables/OCR_table1}
\section{Method}
\label{sec:method}

\subsection{Overall Framework Overview}
Our complete framework of \ourmodel, consisting of the model architecture, as well as the training and inference pipelines, is illustrated in Fig.~\ref{fig:architecture}. We leverage a diffusion model’s strong generative prior for our task, TAIR, incorporating ControlNet~\cite{zhang2023adding} as a conditioning mechanism within the T2I diffusion framework. To handle unknown complex degradations in low-quality (LQ) images, a lightweight degradation removal module is employed following prior work~\cite{lin2024diffbir, ai2024dreamclear}, improving generalization and conditioning reliability. A text-spotting module guides training via its supervisory signals. Unlike conventional methods relying on ResNet features~\cite{liu2020abcnet, yu2020towards, kittenplon2022towards, ye2023deepsolo, zhang2022text}, we utilize semantically rich diffusion features pretrained on large-scale image-text pairs. These diffusion features not only enhance text spotting performance, benefiting from their innate capability to generate text images, but also improve restoration quality through shared representations. The module outputs both localization and recognition of text, with recognized text used as a prompt during diffusion inference to refine textual restoration. A detailed analysis is provided in the supplementary material.

\subsection{Architecture}
\label{sec:method_tair_arch}

\paragraph{Light-weight degradation removal module.}
Given the complexity and entanglement of various degradations in LQ images, using them directly as conditioning signals for image restoration can lead to instability and difficulty in learning text-aware image features. Following prior work~\cite{lin2024diffbir, ai2024dreamclear}, a lightweight degradation removal module~\cite{liang2021swinir} is initially applied to the LQ image to obtain a mildly denoised and smoothed image, thereby slightly reducing noise in text regions. This image then serves as a more reliable condition. It is subsequently encoded by a VAE encoder~\cite{kingma2013auto} to produce the conditioning latent $c$.

\paragraph{Diffusion based image restoration module.}
The image restoration module is based on a diffusion model architecture comprising a U-Net $\mathcal{U}$ and a Control-Net $\mathcal{C}$. The conditioning latent $c$, together with the prompt $p_t$, is processed to guide restoration within the U-Net~\cite{rombach2022high}. Specifically, the high-quality (HQ) image is encoded into a latent representation $z_0$, followed by a diffusion process that progressively adds noise, producing a noisy latent $z_t$ at timestep $t \sim \mathcal{U}(1, T)$, where $\mathcal{U}$ denotes a uniform distribution over diffusion steps 1 to $T$. To enhance conditional learning, the noisy latent and conditioning latent are concatenated as $c_t = \mathrm{concat}(z_t, c)$ and fed into the Control-Net $\mathcal{C}$.

\paragraph{Text-spotting module.}
Incorporating a text-spotting module within the restoration framework enables the diffusion model to learn text-aware image features via gradient supervision. We adopt a transformer-based architecture to leverage the attention mechanism for accurate text detection and recognition. Specifically, an encoder $\mathcal{E}$ and two decoders, $\mathcal{D}^{\text{det}}$ and $\mathcal{D}^{\text{rec}}$, are employed for multi-scale diffusion feature processing, text detection, and text recognition, respectively. The text-spotting module outputs a set of polygon-character tuples defined as
$Y = {(d^{(i)}, r^{(i)})}_{i=1}^K$,
where $i$ indexes each text instance, and $K$ denotes the number of instances with confidence scores exceeding a threshold $T$. Here, $d^{(i)} = (d_1^{(i)}, \ldots, d_N^{(i)})$ represents the coordinates of $N$ control points forming a polygon, and $r^{(i)} = (r_1^{(i)}, \ldots, r_M^{(i)})$ denotes the $M$ recognized characters.

\subsection{Textual prompt guidance}
\input{figs/timestep_inference}
During the denoising process, the text-spotting module outputs polygon-character tuples $\{(d^{(i)}, r^{(i)})\}_{i=1}^{K}$ at each timestep $t$. To facilitate TAIR, a recognition-guided prompt is constructed as $p_t = \mathrm{Prompter}(\{r^{(i)}\}_{i=1}^{K})$, where $\mathrm{Prompter}(\cdot)$ formats the recognized texts for conditioning. This prompt dynamically guides the diffusion model to refine textual content based on intermediate recognition results as shown in Fig.~\ref{fig:timestep_inference}. The prompt $p_t$ is used as input in the denoising step at timestep $t-1$ as $z_{t-1} = \epsilon_\theta(z_t, t, p_t, c_t)$, where $z_t$ is the noisy latent at timestep $t$, $c_t$ is the conditioning signal, and $\epsilon_\theta$ is the noise prediction network. More details can be found in the supplementary material.

%\vspace{-5pt}
\subsection{Training}

%\vspace{-5pt}
\paragraph{Stage 1.}  
Stage 1 training aims to learn an image restoration model that can restore textual content in images, guided by prompts describing the text. During this stage, only the diffusion components comprising U-Net $\mathcal{U}$ and ControlNet $\mathcal{C}$ are optimized, while all other modules remain frozen. Given the diffusion timestep $t$, prompt $p_t$, and control input $c_t$, the model learns a noise prediction network $\epsilon_\theta$ that estimates the noise added to the HQ latent $z_t$. The training loss in Stage1 is defined as:
\begin{equation}
    \mathcal{L}_\text{diff} = \mathbb{E}_{z_0,\, t,\, p_t,\, c_t,\, \epsilon \sim \mathcal{N}(0, 1)} \left[ \left\| \epsilon - \epsilon_\theta(z_t, t, p_t, c_t) \right\|_2^2 \right].
\end{equation}

\paragraph{Stage 2.}
In this stage, only the text-spotting module is trained, while all other components remain frozen. Following transformer-based text-spotting methods~\cite{huang2024bridging, zhang2022text, qiao2024dntextspotter}, detection and recognition losses are computed via bipartite matching~\cite{carion2020end}, solved using the Hungarian algorithm~\cite{kuhn1955hungarian}. Specifically, two separate loss functions are applied to the encoder and the dual decoder:
\begin{align}
\mathcal{L}_{\text{enc}} &= \sum_m \left( \lambda_{\text{cls}}\cdot \mathcal{L}^{(m)}_{\text{cls}} + \lambda_{\text{box}}\cdot \mathcal{L}^{(m)}_{\text{box}} + \lambda_{\text{gIoU}}\cdot \mathcal{L}^{(m)}_{\text{gIoU}} \right), \label{eq:ts_enc} \\
\mathcal{L}_{\text{dec}} &= \sum_n \left( \lambda_{\text{cls}}\cdot \mathcal{L}^{(n)}_{\text{cls}} + \lambda_{\text{poly}}\cdot \mathcal{L}^{(n)}_{\text{poly}} + \lambda_{\text{char}}\cdot \mathcal{L}^{(n)}_{\text{char}} \right), \label{eq:ts_dec}
\end{align}
where $m$ and $n$ denote the numbers of instances with confidence scores exceeding a threshold $T$, and $\mathcal{L}_\text{cls}$, $\mathcal{L}_\text{box}$, $\mathcal{L}_\text{gIoU}$, $\mathcal{L}_\text{poly}$, and $\mathcal{L}_\text{char}$ denote the text classification loss, bounding box regression loss, generalized IoU loss~\cite{rezatofighi2019generalized}, polygon regression loss (L1), and character recognition loss (cross-entropy), respectively. Each loss is weighted by its corresponding factor: $\lambda_{\text{cls}}$, $\lambda_{\text{box}}$, $\lambda_{\text{gIoU}}$, $\lambda_{\text{poly}}$, and $\lambda_{\text{char}}$.

\paragraph{Stage 3.}
In the final training stage, we optimize both the diffusion-based image restoration module and the text spotting module. The total loss function is formulated as:
\begin{equation}
    \mathcal{L} = \mathcal{L}_\text{diff} + \lambda( \mathcal{L}_\text{enc} + \mathcal{L}_\text{dec}), \label{eq:stage3}
\end{equation}
where $\lambda$ is a weight value.

%% file: figs/architecture.tex
\begin{figure}[t]
    \centering
    \includegraphics[width=\linewidth]{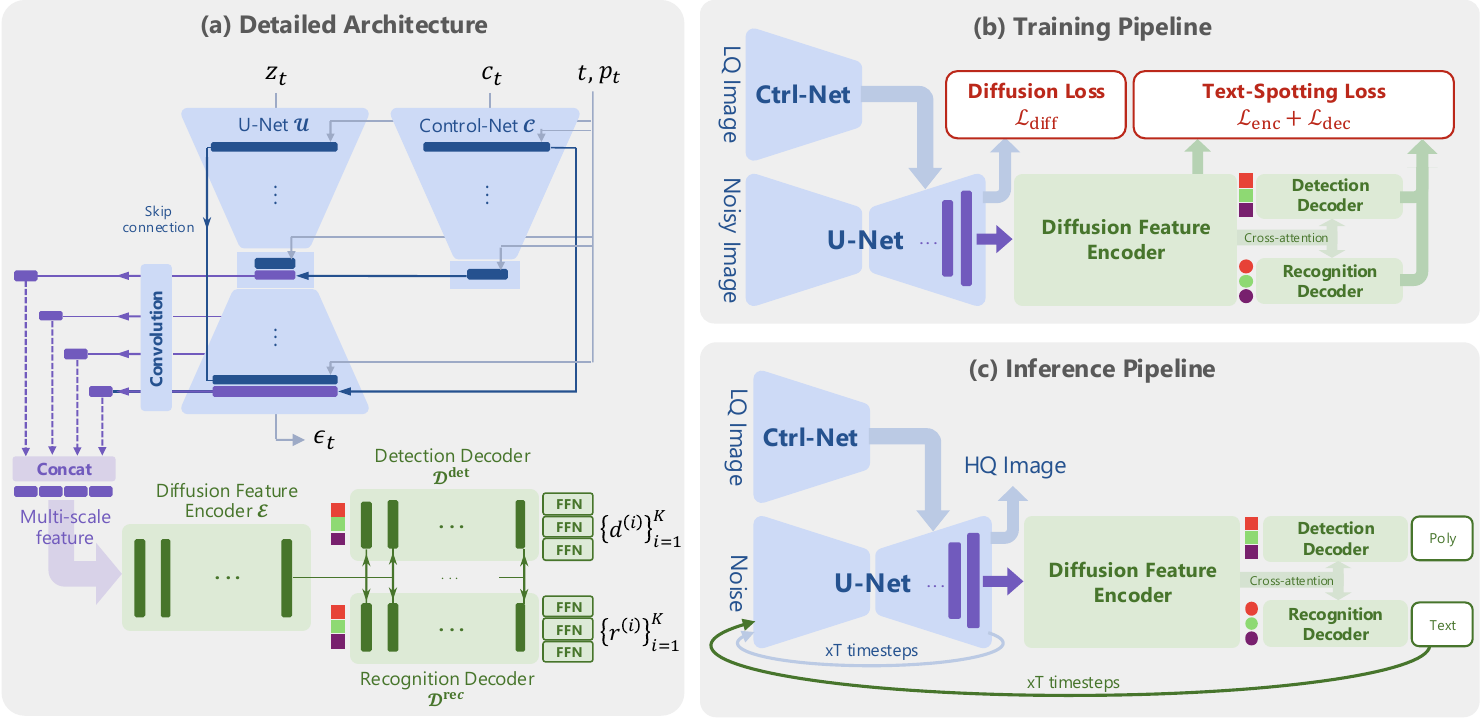}
    \caption{
    \textbf{Overview of the \ourmodel{} architecture, training, and inference pipeline.} TAIR integrates a text-spotting module into a diffusion image restoration framework, using text supervision during training and recognized text as a prompt at inference to enhance text-aware image restoration.}
    \vspace{-10pt}
    \label{fig:architecture}

\end{figure}

%% file: tables/OCR_table1.tex
\begin{table}[t]
    \centering
    \resizebox{\textwidth}{!}{
        \begin{tabular}{ c|l|ccccc|ccccc }
            \toprule
            \multirow{3}{*}{\textbf{Deg. Level}} &
            \multirow{3}{*}{\textbf{Model}} &
            \multicolumn{5}{c|}{\textbf{ABCNet v2~\cite{liu2021abcnet}}} &
            \multicolumn{5}{c}{\textbf{TESTR~\cite{zhang2022text}}} \\ 
            
            \cmidrule(lr){3-7}
            \cmidrule(lr){8-12}
            & & 
            \multicolumn{3}{c}{Detection} &
            \multicolumn{2}{c|}{End-to-End} &
            \multicolumn{3}{c}{Detection} &
            \multicolumn{2}{c}{End-to-End} \\
            
            \cmidrule(lr){3-5}
            \cmidrule(lr){6-7}
            \cmidrule(lr){8-10}
            \cmidrule(lr){11-12}
            & & Precision($\uparrow$) & Recall($\uparrow$) & F1-Score($\uparrow$) & None($\uparrow$) & Full($\uparrow$) & Precision($\uparrow$) & Recall($\uparrow$) & F1-Score($\uparrow$) & None($\uparrow$) & Full($\uparrow$) \\ 
            \midrule
            % ---------- HQ (Top) ----------
            - & HQ (GT) & 92.16 & 86.85 & 89.43 & 71.79 & 82.05 & 92.59 & 87.81 & 90.13 & 75.90 & 84.18 \\ 
            \midrule
            
            % ---------- Level 1 ----------
            \multirow{10}{*}{Level1} 
            & LQ (Lv1) & 89.79 & 29.51 & 44.42 & 24.29 & 34.25 & 84.01 & 30.24 & 44.47 & 25.93 & 34.73 \\ 
            \cmidrule{2-12}
            & Real-ESRGAN~\cite{wang2021real} & 83.79 & 43.34 & 57.13 & 21.45 & 30.12 & 85.19 & 41.98 & 56.24 & 22.90 & 31.52 \\
            & SwinIR~\cite{liang2021swinir} & \underline{84.95} & 40.93 & 55.25 & 22.70 & 31.26 & \underline{87.93} & 39.62 & 54.63 & 25.50 & 33.75 \\ 
            & ResShift~\cite{yue2023resshift} & 81.93 & 40.07 & 53.82 & 20.40 & 28.74 & \textbf{88.47} & 35.81 & 50.98 & 22.39 & 30.14 \\
            & StableSR~\cite{wang2024exploiting} & 77.90 & 55.44 & 64.78 & 21.93 & 29.29 & 84.44 & \underline{50.68} & 63.34 & 24.02 & 31.84 \\ 
            & DiffBIR~\cite{lin2024diffbir} & 76.29 & 56.44 & 64.88 & \underline{23.14} & \underline{32.73} & 84.00 & 52.13 & 64.34 & \underline{25.51} & \underline{35.47} \\
            & SeeSR~\cite{wu2024seesr} & 70.00 & \textbf{61.88} & \underline{65.69} & 20.16 & 28.63 & 78.85 & \textbf{55.76} & \underline{65.32} & 23.31 & 32.82 \\
            & SUPIR~\cite{yu2024scaling} & 43.64 & 49.46 & 46.37 & 14.58 & 19.34 & 53.02 & 46.19 & 49.37 & 17.44 & 22.29 \\

            & FaithDiff~\cite{chen2024faithdiff} & 69.16 & 61.51 & 65.12 & 20.44 & 27.78 & 78.80 & 57.12 & 66.23 & 22.50 & 31.59 \\ 
            
            \cmidrule{2-12} 
            & \textbf{\ourmodel{} (Ours)} & \textbf{85.29} & \underline{58.34} & \textbf{69.29} & \textbf{26.59} & \textbf{35.69} & 87.50 & \underline{54.90} & \textbf{67.47} & \textbf{28.19} & \textbf{36.99} \\ 
            \midrule
            \midrule

            % ---------- Level 2 ----------
            \multirow{10}{*}{Level2} 
            & LQ (Lv2) & 87.67 & 22.89 & 36.30 & 20.49 & 27.82 & 78.45 & 23.93 & 36.68 & 20.49 & 27.37 \\ 
            \cmidrule{2-12}
            & Real-ESRGAN~\cite{wang2021real} & \underline{81.42} & 41.12 & 54.64 & 18.31 & 24.88 & 84.92 & 38.80 & 53.27 & 19.29 & 27.50 \\ 
            & SwinIR~\cite{liang2021swinir} & 80.14 & 37.31 & 50.91 & 17.82 & 24.93 & \underline{85.43} & 34.81 & 49.47 & 19.07 & 26.99 \\ 
            & ResShift~\cite{yue2023resshift} & 81.11 & 35.22 & 49.12 & 17.89 & 25.54 & 85.18 & 32.05 & 46.57 & 17.26 & 26.09 \\ 
            & StableSR~\cite{wang2024exploiting} & 75.49 & 51.95 & 61.55 & 19.55 & 26.69 & 79.03 & 48.87 & 60.39 & 20.06 & 27.68 \\ 
            & DiffBIR~\cite{lin2024diffbir} & 72.96 & 53.94 & 62.03 & \underline{19.60} & \underline{27.52} & 79.69 & 50.50 & 61.82 & 21.64 & \underline{30.36} \\ 
            & SeeSR~\cite{wu2024seesr} & 68.93 & \textbf{60.65} & \underline{64.53} & 19.48 & 26.72 & 77.50 & \textbf{54.49} & \underline{63.99} & \underline{21.83} & 29.17 \\ 
            & SUPIR~\cite{yu2024scaling} & 42.01 & 45.65 & 43.75 & 13.21 & 17.42 & 53.54 & 43.25 & 47.84 & 15.50 & 19.96 \\ 

            & FaithDiff~\cite{chen2024faithdiff} & 66.62 & 59.34 & 62.77 & 18.94 & 25.99 & 76.16 & 54.17 & 63.31 & 20.98 & 28.56 \\ 
            
            \cmidrule{2-12}
            & \textbf{\ourmodel{} (Ours)} & \textbf{83.02} & \underline{56.30} & \textbf{67.10} & \textbf{24.42} & \textbf{33.23} & \textbf{86.95} & \underline{52.86} & \textbf{65.75} & \textbf{26.39} & \textbf{35.13} \\ 
            \midrule
            \midrule

            % ---------- Level 3 ----------
            \multirow{10}{*}{Level3} 
            & LQ (Lv3) & 85.38 & 13.24 & 22.92 & 12.17 & 16.95 & 76.01 & 15.37 & 25.57 & 12.52 & 17.72 \\ 
            \cmidrule{2-12}
            & Real-ESRGAN~\cite{wang2021real} & \underline{72.48} & 28.65 & 41.07 & 11.89 & 16.37 & 76.51 & 27.02 & 39.93 & 12.13 & 18.22 \\
            & SwinIR~\cite{liang2021swinir} & 74.41 & 25.57 & 38.06 & 11.27 & 16.33 & \underline{78.38} & 24.16 & 36.94 & 11.85 & 17.12 \\ 
            & ResShift~\cite{yue2023resshift} & 75.00 & 22.57 & 34.70 & 10.80 & 15.19 & 81.10 & 20.04 & 32.13 & 9.89 & 15.63 \\
            & StableSR~\cite{wang2024exploiting} & 67.63 & 38.08 & 48.72 & 13.34 & 18.91 & 72.21 & 35.22 & 47.35 & 13.65 & 19.68 \\ 
            & DiffBIR~\cite{lin2024diffbir} & 59.30 & 42.20 & 49.31 & \underline{13.88} & \underline{19.39} & 72.27 & 38.98 & 50.65 & 15.61 & \underline{22.67} \\
            & SeeSR~\cite{wu2024seesr} & 55.06 & \textbf{46.83} & \underline{50.61} & 13.38 & 18.47 & 64.95 & \textbf{43.93} & \underline{52.41} & \underline{14.93} & 20.88 \\ 
            & SUPIR~\cite{yu2024scaling} & 31.05 & 34.72 & 32.78 & 9.07 & 11.77 & 40.78 & 32.77 & 36.34 & 11.21 & 14.02 \\

            & FaithDiff~\cite{chen2024faithdiff} & 56.04 & 47.91 & 51.66 & 13.69 & 19.01 & 69.44 & 45.01 & 54.62 & 15.40 & 21.18 \\

            \cmidrule{2-12} 
            & \textbf{\ourmodel{} (Ours)} & \textbf{81.76} & \underline{44.11} & \textbf{57.30} & \textbf{19.61} & \textbf{27.50} & \textbf{84.50} & \underline{42.02} & \textbf{56.13} & \textbf{19.92} & \textbf{28.34} \\ 

            \bottomrule
        \end{tabular}
    }
    \vspace{5pt}
    \caption{\textbf{Quantitative results of text spotting on \ourdataset.} Each block shows the performance of various image restoration methods under different degradation strengths, evaluated using two text spotting models. ‘None’ refers to recognition without a lexicon, and ‘Full’ denotes recognition with a full lexicon. Best results are in \textbf{bold} and second-best are \underline{underlined}. } 
    \label{tab:ocr_main_table}
\end{table}

%% file: figs/timestep_inference.tex
\begin{figure}[h]
    \includegraphics[width=\linewidth]{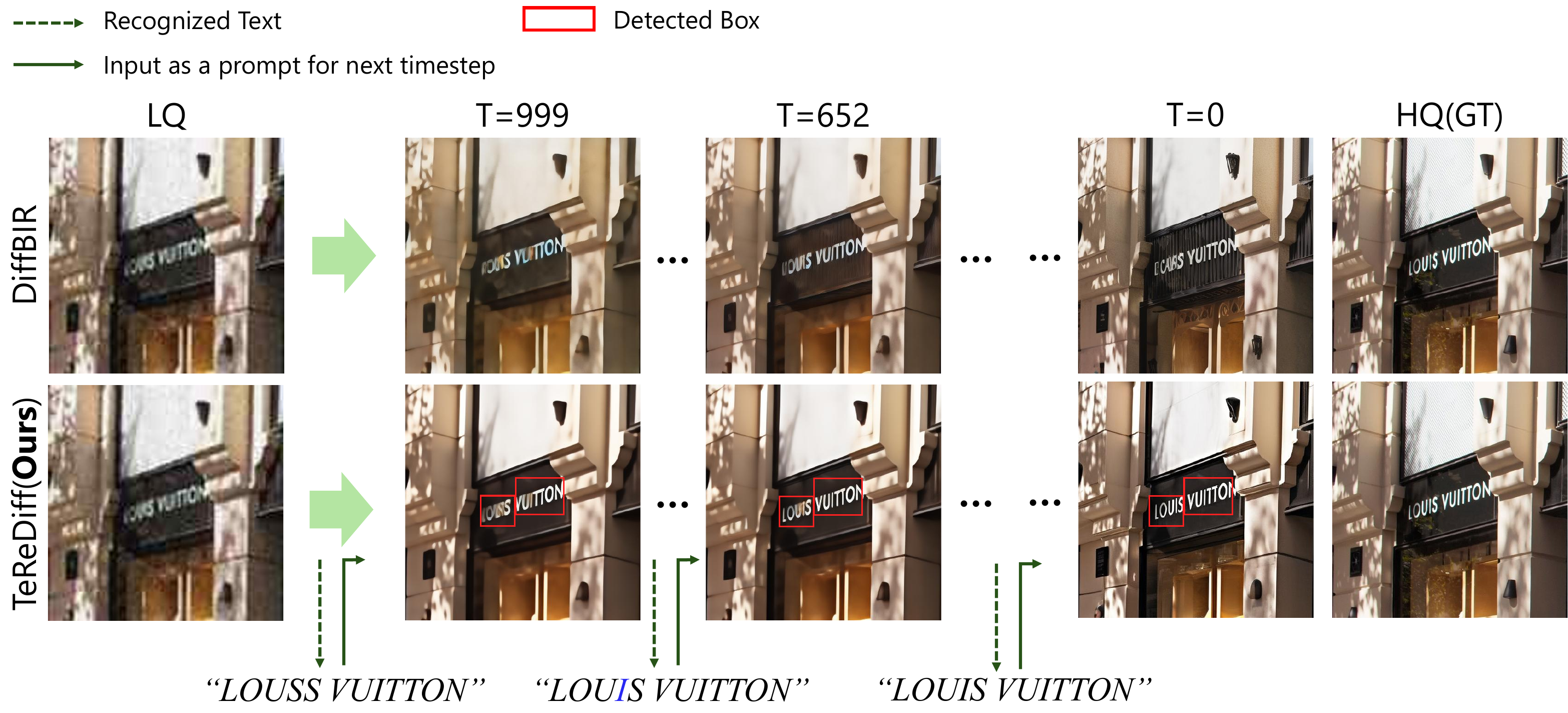}
    \vspace{-10pt}
    \caption{
    \textbf{Example of textual prompt guidance for each timestep.} The outputs of the text-spotting module serve as input prompts for the next timestep. The detected boxes and recognized texts may also change as the image is gradually restored over timesteps.}
    \label{fig:timestep_inference}
\end{figure}

%% file: sections/5_experiment.tex
\input{tables/OCR_table2}

\input{tables/SR_table}

\section{Experiments}

\subsection{Experimental Settings}
\label{sub:setting}
\paragraph{Training and evaluation dataset.}
We train our \ourmodel{} on \ourdataset{}, which contains 100K high-quality 512$\times$512 images. Synthetic degradations are applied using the Real-ESRGAN pipeline \cite{wang2021real}, a method widely adopted in image restoration studies~\cite{yue2023resshift, wang2024exploiting, lin2024diffbir, wu2024seesr, yu2024scaling}. For testing, we curate a separate 1K subset of \ourdataset{}. Similar to DASR~\cite{liang2023dasr} and FaithDiff~\cite{chen2024faithdiff}, we apply three progressively challenging degradation levels to our test set, to evaluate the model in various difficulty settings. We further construct Real-Text, a set of real-world HR–LR pairs extracted from RealSR~\cite{cai2019toward} and DRealSR\cite{wei2020component} with text annotations using our dataset curation pipeline.

\paragraph{Evaluation metrics.}
To evaluate the text restoration performance, we employ off-the-shelf pre-trained text spotting models~\cite{liu2021abcnet,zhang2022text} to perform text detection and recognition on the restored images. We measure Precision, Recall, and F1-scores for text detection results, and F1-scores for end-to-end text recognition results, which are standard in text spotting tasks. We evaluate image restoration performance using both reference and non-reference metrics. For reference-based evaluation, we adopt PSNR and SSIM~\cite{wang2004image} to measure fidelity, and LPIPS~\cite{zhang2018unreasonable} and DISTS~\cite{ding2020image} to assess perceptual quality. Additionally, we use FID~\cite{heusel2017gans} to evaluate the distributional similarity between the restored and ground-truth image sets.
For non-reference evaluation, we report NIQE~\cite{zhang2015feature}, MANIQA~\cite{yang2022maniqa}, MUSIQ~\cite{ke2021musiq}, and CLIPIQA~\cite{wang2023exploring} scores.

\paragraph{Implementation details.} 
\label{exp:impl}
We utilize SwinIR~\cite{liang2021swinir}, DiffBIR~\cite{lin2024diffbir}, SD2.1~\cite{rombach2022high}, and TESTR~\cite{zhang2022text} as the light-weight degradation removal module, Control-Net, UNet, and text-spotting architecture, respectively. The training is performed using the AdamW optimizer with default parameters. The learning rate is set to $1 \times 10^{-4}$ for Stage 1 and Stage 2, and $1 \times 10^{-5}$ for Stage 3. Publicly available checkpoints are used for all models. The input LR and output HR images are both $512 \times 512$ in size. Further details can be found in the supplementary materials.

\subsection{Main Results}
We compare our \ourmodel{} with GAN-based models (Real-ESRGAN~\cite{wang2021real} and SwinIR~\cite{liang2021swinir}) and diffusion-based models (ResShift~\cite{yue2023resshift}, StableSR~\cite{wang2024exploiting}, DiffBIR~\cite{lin2024diffbir}, SeeSR~\cite{wu2024seesr}, SUPIR~\cite{yu2024scaling}, and FaithDiff~\cite{chen2024faithdiff}). More quantitative and qualitative results are provided in supplementary materials.

\paragraph{Quantitative comparisons.}
Tab.~\ref{tab:ocr_main_table} shows detection and recognition metrics on \ourdataset{}. Our \ourmodel{} achieves the best F1-score at every level on both text spotting models. Prior models often lose recognition accuracy at Level2 and Level3, dropping below the raw low-resolution inputs because stronger degradations intensify \textit{text-image hallucination}. In contrast, TAIR consistently restores textual regions, preserving recognition performance even under the strongest degradations. Furthermore, Tab.~\ref{tab:real_ocr_table} illustrates the performance of our \ourmodel{} in real-world scenarios. For image quality, Tab.~\ref{tab:sr_main_table} demonstrates that our model outperforms the baseline on reference-based metrics while achieving comparable performance on non-reference metrics. These results indicate that our model restores text regions effectively without compromising overall image restoration performance.

\paragraph{Qualitative comparison.}
Fig.~\ref{fig:teaser} shows representative results on SA-Text test set. Previous restoration methods often produce blurry characters, inconsistent stroke widths, which are text–image hallucinations under severe degradations due to reliance on the generative priors of diffusion models. In contrast, \ourmodel{} consistently restores readable text in challenging regions. These qualitative improvements align closely with the quantitative results in Tab.~\ref{tab:real_ocr_table}, highlighting our model's effectiveness in enhancing text clarity without compromising overall image quality.

\input{tables/abl_table}
\subsection{Ablation Study}
% ablation#1. 

We conduct ablation studies on SA-Text (Level 2) to evaluate the effectiveness of multi-stage training, the importance of prompting textual content for text-aware image restoration, and to compare different prompting styles. Additional ablations on all levels of SA-Text and on Real-Text are provided in the supplementary material.

\paragraph{Effectiveness of multi-stage training.} 
To assess the effectiveness of multi-stage training for text-aware restoration, we compare results from models trained up to Stage 1 and Stage 3, as shown in Tab.\ref{tab:abl}\subref{tab:ablations_two_tables_a}. Advancing training to Stage 3 consistently improves performance across all caption types. The $\mathrm{null}$ type refers to cases where no prompt is provided during inference. The $\mathrm{pr}$ type uses a prompt sourced from either an external prompter\cite{liu2023visual} or our internal text-spotting module. The $\mathrm{gt}$ type uses the ground-truth text present in the LQ image.

\paragraph{Importance of prompting for text restoration.} 
Performance comparisons among the $\mathrm{null}$, $\mathrm{pr}$, and $\mathrm{gt}$ settings in Tab.~\ref{tab:abl}\subref{tab:ablations_two_tables_a} reveal that prompting with textual content from the LQ image enhances restoration quality in both Stage 1 and Stage 3, highlighting the importance of text prompts for improved text-aware restoration. In particular, using ground-truth text prompts results in substantial improvements in both stages. The small performance gap between the two stages under the gt setting suggests that supplying the restoration module with accurate ground-truth text constitutes an ideal scenario, effectively defining an upper bound for text-aware restoration performance.

\paragraph{Text prompting style.}
Tab.~\ref{tab:abl}\subref{tab:ablations_two_tables_b} shows that the $\mathrm{tag}$ and $\mathrm{cap}$ prompting styles yield similar performance with ground-truth (gt) text. However, when using predicted text from the text-spotting module, a notable performance gap arises. This highlights the importance of prompting styles for predicted textual content. Given texts “text1,” “text2,” and “text3,” the caption style (subscript $\mathrm{cap}$) uses the format: “A realistic scene where the texts {text1, text2, ...} appear clearly on signs, boards, buildings, or other objects.” The tag style (subscript $\mathrm{tag}$) uses: “{text1, text2, ...}.”

%% file: tables/OCR_table2.tex
\begin{table}[t]
    \centering
    \resizebox{0.9\textwidth}{!}{
        \begin{tabular}{ l|ccccc|ccccc } 
            \toprule
            \multirow{3}{*}{\textbf{Model}} &
            \multicolumn{5}{c|}{\textbf{ABCNet v2~\cite{liu2021abcnet}}} & 
            \multicolumn{5}{c}{\textbf{TESTR~\cite{zhang2022text}}} \\ 
    
            \cmidrule(lr){2-6}
            \cmidrule(lr){7-11}
            &
            \multicolumn{3}{c}{Detection} &
            \multicolumn{2}{c|}{End-to-End} &
            \multicolumn{3}{c}{Detection} &
            \multicolumn{2}{c}{End-to-End} \\

            \cmidrule(lr){2-4}
            \cmidrule(lr){5-6}
            \cmidrule(lr){7-9}
            \cmidrule(lr){10-11}
          
            & Precision($\uparrow$) & Recall($\uparrow$) & F1-Score($\uparrow$) & None($\uparrow$) & Full($\uparrow$) & Precision($\uparrow$) & Recall($\uparrow$) & F1-Score($\uparrow$) & None($\uparrow$) & Full($\uparrow$) \\ 
            \midrule 
            
            HQ (GT) & 90.03 & 85.52 & 87.72 & 72.06 & 79.48 & 90.29 & 85.77 & 87.97 & 74.50 & 81.72 \\ 
            \cmidrule(lr){1-11} 
            LQ & 89.10 & 44.97 & 59.77 & 42.64 & 50.21 & 85.33 & 51.61 & 64.32 & 47.08 & 55.11 \\ 
            \cmidrule(lr){1-11} 
            Real-ESRGAN~\cite{wang2021real} & 79.15 & 52.70 & 63.27 & 35.30 & 39.88 & \underline{82.67} & 53.94 & 65.29 & 38.16 & 42.36 \\ 
            SwinIR~\cite{liang2021swinir} & 80.29 & 47.45 & 59.64 & 38.39 & 42.63 & \underline{82.92} & 47.89 & 60.72 & 39.97 & 44.56 \\ 
            ResShift~\cite{yue2023resshift} & \underline{81.17} & 33.76 & 47.69 & 30.95 & 34.87 & 82.23 & 39.91 & 53.74 & 35.31 & 39.99 \\ 
            StableSR~\cite{wang2024exploiting} & 79.79 & 59.89 & \underline{68.42} & \underline{41.23} & \underline{47.64} & 82.19 & 60.39 & 69.62 & \underline{42.53} & \underline{49.39} \\ 
            DiffBIR~\cite{lin2024diffbir} & 66.04 & 59.69 & 62.71 & 33.75 & 40.05 & 76.33 & 61.87 & 68.35 & 39.27 & 46.11 \\ 
            SeeSR~\cite{wu2024seesr} & 68.12 & 63.46 & 65.71 & 37.11 & 43.43 & 74.29 & 62.47 & 67.87 & 40.34 & 46.54 \\ 
            SUPIR~\cite{yu2024scaling} & 44.00 & 40.56 & 42.21 & 22.29 & 25.03 & 53.08 & 44.47 & 48.39 & 27.25 & 30.59 \\ 
            FaithDiff~\cite{chen2024faithdiff} & 71.21 & \underline{64.50} & 67.69 & 38.81 & 44.28 & 76.90 & \underline{65.20} & \underline{70.57} & 41.64 & 47.97 \\
            \cmidrule(lr){1-11} 
            \textbf{\ourmodel{} (Ours)} & \textbf{83.95} & \textbf{67.58} & \textbf{74.88} & \textbf{48.39} & \textbf{55.01} & \textbf{84.30} & \textbf{67.37} & \textbf{74.89} & \textbf{49.39} & \textbf{56.45} \\ 
            \bottomrule    
        \end{tabular}
    }
    \vspace{5pt}
    \caption{\textbf{Quantitative results of text spotting on Real-Text.} Using text spotting models, we evaluate the accuracy of text detection and recognition of the restored images on Real-Text.}
    \label{tab:real_ocr_table}    
\end{table}

%% file: tables/SR_table.tex
\begin{table}[t]
    \centering
    \resizebox{\textwidth}{!}{
        \begin{tabular}{ l|c|ccccccccc } 
        
            \toprule
            \addlinespace[4pt]
            Dataset &
            Model &
            PSNR($\uparrow$) &
            SSIM($\uparrow$)&
            LPIPS($\downarrow$) &
            DISTS($\downarrow$) &
            FID($\downarrow$) &
            NIQE($\downarrow$) &
            MANIQA($\uparrow$) &
            MUSIQ($\uparrow$) &
            CLIPIQA($\uparrow$)
            \\
            \addlinespace[3pt]

            \midrule 
            \multirow{3}{*}{$\text{SA-Text}_\text{ test}$}

            & DiffBIR & 
            \underline{19.58} & \underline{0.4965} & \underline{0.3636} & \underline{0.2080} & \underline{45.10} & \textbf{5.107} & \underline{0.6771} & \textbf{73.33} & \textbf{0.6589} \\
  
            & DiffBIR$\dagger$ & 
            16.81 & 0.4638 & 0.4001 & 0.2219 & 47.28 & \underline{5.449} & \textbf{0.6890} & \underline{72.55} & \underline{0.6345} \\
            
            & \textbf{\ourmodel{} (Ours)} & 
            \textbf{19.71} & \textbf{0.5717} & \textbf{0.2828} & \textbf{0.1702} & \textbf{36.94} & 5.452 & 0.6471 & 72.07 & 0.6145 \\

            \midrule 
            \multirow{3}{*}{Real-Text} 
            
            & DiffBIR & 
            \underline{23.00} & \underline{0.6516} & \underline{0.4108} & \underline{0.2925} & 87.46 & \textbf{7.054} & \textbf{0.6147} & \textbf{66.84} & \textbf{0.5679}  \\
            
            & DiffBIR$\dagger$ & 
            19.11 & 0.5107 & 0.5233 & 0.3127 &\underline{80.59} & 8.42 & \underline{0.5873} & 61.46 & \underline{0.4961}  \\
            
            & \textbf{\ourmodel{} (Ours)} & 
            \textbf{23.37} & \textbf{0.7849} & \textbf{0.2848} & \textbf{0.2386} & \textbf{68.94} & \underline{7.643} & 0.5637 & \underline{62.02} & 0.4545 \\
        
        \bottomrule    
        \end{tabular}
    }
\vspace{3pt}
\caption{\textbf{Quantitative results of image restoration.} We evaluate the image quality of our \ourmodel{} compared with its baseline. The degradation pipeline~\cite{wang2021real} used in prior works is applied. DiffBIR~\cite{lin2024diffbir}$\dagger$ (denoted as v2.1 in Fig.~\ref{fig:teaser}) uses GitHub-released weights that are further trained from those in the original paper.}
\label{tab:sr_main_table}    
\end{table}

%% file: tables/abl_table.tex
\begin{table*}[t]
\centering
\subfloat[
\textbf{Multi-stage training and textual prompting.} Stage-wise training and textual prompting of the restoration module enhance text restoration.
\label{tab:ablations_two_tables_a}
]{
\begin{minipage}{0.45\linewidth}
\centering
% \small
\tablestyle{5pt}{1.0}
\resizebox{\textwidth}{!}{
\begin{tabular}{lccccc}
\toprule
\multirow{3}{*}{\textbf{Model}} & \multicolumn{5}{c}{\textbf{TESTR~\cite{zhang2022text}}} \\
\cmidrule(lr){2-4}
\cmidrule(lr){5-6}

&
\multicolumn{3}{c}{Detection} &
\multicolumn{2}{c}{End-to-End} \\

\cmidrule(lr){2-4}
\cmidrule(lr){5-6}
 & Precision($\uparrow$) & Recall($\uparrow$) & F1-Score($\uparrow$) & None($\uparrow$) & Full($\uparrow$) \\
\midrule
Stage1\textsubscript{$\mathrm{null}$} & 81.77 & 47.37 & 59.99 & 21.24 & 29.79  \\
Stage1\textsubscript{$\mathrm{pr}$} & \underline{82.01} & \underline{49.82} & \underline{61.99} & \underline{24.76} & \underline{31.70}  \\
Stage1\textsubscript{$\mathrm{gt}$} & \textbf{85.09} & \textbf{61.56} & \textbf{71.44} & \textbf{32.51} & \textbf{42.71} \\
\midrule
Stage3\textsubscript{$\mathrm{null}$} & 84.47 & \underline{56.21} & \underline{67.50} & 23.46 & 32.72 \\
Stage3\textsubscript{$\mathrm{pr}$} & \textbf{86.95} & 52.86 & 65.75 & \underline{26.39} & \underline{35.13}  \\
Stage3\textsubscript{$\mathrm{gt}$} & \underline{86.18} & \textbf{61.60} & \textbf{71.85} & \textbf{33.31} & \textbf{43.40} \\
\bottomrule
\end{tabular}}
\end{minipage}
}
\hfill
\subfloat[
\textbf{Prompting styles for text restoration.} The choice of prompting style impacts restoration performance more noticeably with predicted text than with ground-truth text.
\label{tab:ablations_two_tables_b}
]{
\begin{minipage}{0.45\linewidth}
\centering
% \small
\tablestyle{5pt}{1.0}
\resizebox{\textwidth}{!}{
\begin{tabular}{lccccc}
\toprule
\multirow{2}{*}{\textbf{Caption}} & \multicolumn{5}{c}{\textbf{TESTR~\cite{zhang2022text}}} \\
\cmidrule(lr){2-4}
\cmidrule(lr){5-6}

&
\multicolumn{3}{c}{Detection} &
\multicolumn{2}{c}{End-to-End} \\
\cmidrule(lr){2-4}
\cmidrule(lr){5-6}
 & Precision($\uparrow$) & Recall($\uparrow$) & F1-Score($\uparrow$) & None($\uparrow$) & Full($\uparrow$) \\
\midrule
Pred\textsubscript{$\mathrm{tag}$}     & \underline{83.25} & \textbf{59.25} & \textbf{69.23} & \underline{24.26} & \underline{31.94} \\
Pred\textsubscript{$\mathrm{cap}$}     & \textbf{86.95} & \underline{52.86} & \underline{65.75} & \textbf{26.39} & \textbf{35.13}  \\
\midrule
GT\textsubscript{$\mathrm{tag}$}         & \underline{84.40} & \textbf{62.56} & \textbf{71.86} & \underline{32.02} & \underline{42.12} \\
GT\textsubscript{$\mathrm{cap}$}         & \textbf{86.18} & \underline{61.60} & \underline{71.85} & \textbf{33.31} & \textbf{43.40} \\
\bottomrule
\end{tabular}}
\end{minipage}
}
\vspace{2pt}
\captionsetup{justification=raggedright,singlelinecheck=false}
\caption{\textbf{Ablation study conducted on SA-Text (level 2 degradation).} Subscripts $\mathrm{null}$, $\mathrm{pr}$, and $\mathrm{gt}$ indicate the use of a null prompt, a prompt generated by a prompter, and a ground-truth prompt, respectively. Stage1 and Stage3 refer to models trained until Stage 1 and Stage 3, respectively. The $\mathrm{pr}$ prompt is generated by a LLaVA captioner~\cite{liu2023visual} in Stage1 and by our text-spotting module in Stage3. Lastly, $\mathrm{tag}$ and $\mathrm{cap}$ denote different prompting styles.}
\label{tab:abl}
\end{table*}

%% file: sections/6_conclusion.tex
\section{Conclusion and Future Work}
\label{sec:conclusion}
We revisit image restoration with a new focus: Text-Aware Image Restoration (TAIR), which targets the recovery of textual content in degraded images. This area remains largely unexplored due to the absence of large-scale, annotated datasets. To address this gap, we introduce SA-Text, a curated dataset that uses VLMs to provide automated supervision for text restoration. Models trained on SA-Text show significantly better performance than existing methods. Despite these advances, TAIR still faces important challenges. Performance declines noticeably with small text, where even the slightest degradation greatly reduces legibility. Text instances in complex natural scenes are also difficult to detect and recognize due to visual clutter and diverse layouts. Future research directions include incorporating more diverse real-world data, enhancing supervision quality, and investigating advanced prompting techniques. We hope that TAIR and SA-Text will inspire further research combining text understanding and image restoration.

%% file: sections/supple.tex
\appendix
%%%%%%%%%%%%%%%%%%%%%%%%%%%%%%%%%%%%%%%%%%%%%%%%%%%%%%%%%%%%

\newpage

%\begin{center}
%    \textbf{\Large Text-Aware Image Restoration with Diffusion Models} \\
%    \vspace{5pt}
%    \textbf{\large - Supplementary Material - }
%    \vspace{5pt}
%\end{center}
\section*{\Large Appendix}

This supplementary material is organized as follows. Sec.\ref{sup:A_extended} provides additional details on the SA-Text curation pipeline, focusing on the handling of false negatives and incorrect detections. Sec.\ref{sup:B_diff_feat} demonstrates the effectiveness of leveraging diffusion features for training the text spotting module. Sec.\ref{sup:C_impl_detail} describes the implementation of the overall TAIR framework, including the evaluation metrics for text restoration, model architecture, training stages, and the text-spotting module used as a prompter. Sec.\ref{sup:D_add_quan} presents extended quantitative results, including comparisons with baselines across three levels of degradation on SA-Text, and the Real-Text dataset, which captures real-world degradation scenarios. Furthermore, we report the results of a user study conducted to measure human preference on the restoration results. Finally, Sec.~\ref{sup:E_add_qual} provides qualitative comparisons of our \ourmodel{} against conventional image restoration methods.

\input{supple/sec/A_sa_text}

\input{supple/sec/B_diff_feature}

\input{supple/sec/C_impl_detail}

\input{supple/sec/D_add_quan}

\input{supple/sec/E_add_qual}
% \input{supple/sec/F_error}

%% file: supple/sec/A_sa_text.tex
\section{SA-Text Curation Pipeline}
\label{sup:A_extended}
% \input{figs/pipeline_results}
% \subsection{Detection and Patch Cropping Details}
% In this section, we provide additional details on the dataset curation pipeline. 
We apply our dataset curation pipeline to a subset of the SA-1B~\cite{kirillov2023segment} dataset, a large-scale dataset originally designed for segmentation tasks. 
SA-1B~\cite{kirillov2023segment} consists of 11M high-resolution images ($3,300\times4,950$ pixels on average) that have been downsampled so that their shortest side is $1,500$ pixels. 
This meets our requirements for high-quality images with sufficient resolution, suitable for the image restoration task. 
After processing with our dataset curation pipeline on a subset of SA-1B~\cite{kirillov2023segment} (18.6\% of the dataset), we are able to curate 100K high-quality crops that are densely annotated with text instances, enabling scalable dataset creation specifically for TAIR.
We expect that using our pipeline on other high-quality large-scale datasets can provide additional data if required.

For the dedicated detection model used in the patch and box cropping stages, we utilize a state-of-the-art text spotting model, DG-Bridge Spotter~\cite{huang2024bridging}. 
As shown in Fig.~\ref{fig:pipeline_results}, text instances initially missed by the detection model (false negatives) on full images are subsequently captured when the model is applied to smaller crops. 
We observe that running the detection model on smaller crops occasionally results in false positives, defined as detections of text in areas without actual text. 
Nonetheless, these instances are reliably filtered out in the subsequent VLM filtering stage, as at least one of the VLMs consistently identifies regions without text.
Despite the strong recognition capabilities of DG-Bridge Spotter~\cite{huang2024bridging}, we opt for VLMs due to their consistently superior recognition accuracy.

For filtering out out-of-focus and intentionally blurred images, we employ a VLM to classify each image's blurriness. 
We find that prompting the VLM to perform multiclass classification (as illustrated in Fig.~\ref{fig:data_pipeline}) offers improved granularity and stricter filtering compared to using a simpler binary classification prompt ("blurry" vs. "not blurry"). 
Examples are shown in Fig. \ref{fig:blur_levels}. 
After classification, crops labeled as Levels 1 and 2 (very blurry and slightly blurry, respectively) are filtered out from the final dataset. This VLM-based blur filtering ensures that only high-quality, sharply focused crops are used for TAIR.
\input{figs/blur_levels}
% Although the blurrier images are unsuitable for the TAIR task, we suspect that they have value as data for the text spotting task, as they are still densely annotated for text. 
% Thus, we release the full set of annotated images, with the relevant "blurriness" tags for reference. The number of images corresponding to each level is outlined in Tab. \ref{}.  

%% file: figs/blur_levels.tex
\begin{figure}[t]
    \centering
    \includegraphics[width=\linewidth]{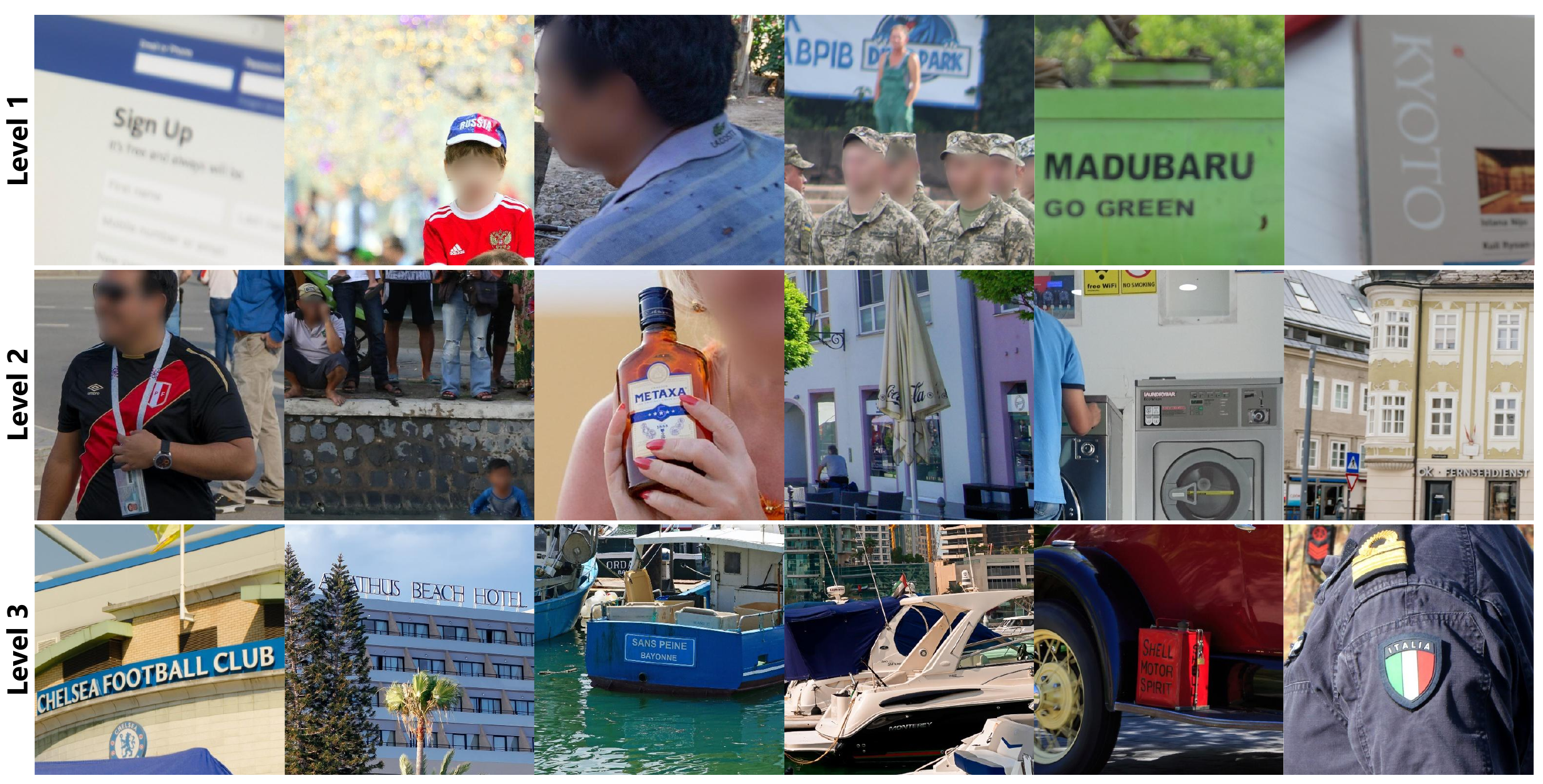}
    \vspace{-15pt}
    \caption{
    \textbf{Examples of images classified by blurriness.}  Images are categorized into Levels 1–3: Level 1 (very blurry), Level 2 (slightly blurry), and Level 3 (clearly focused). Images classified as Level 1 and Level 2 are excluded from the final dataset to ensure that only clearly focused (Level 3) images are used for TAIR.}
    \label{fig:blur_levels}
\end{figure}

%% file: supple/sec/B_diff_feature.tex
\section{Diffusion Features for Text Spotting}
\label{sup:B_diff_feat}
\input{supple/tab/sup_diff_feat}
As mentioned in Sec.~\ref {sec:method}, we demonstrate that a text spotting module can be trained directly using diffusion features for the text spotting task~\cite{zhang2022text, ye2023deepsolo, qiao2024dntextspotter, huang2024bridging} instead of the commonly used ResNet~\cite{he2016deep} backbone, and that this approach not only enables effective training but also yields superior performance. To the best of our knowledge, this is the first work to demonstrate that diffusion features are suitable for effectively learning text spotting.

\subsection{Experimental Details} 
We train TESTR~\cite{zhang2022text} from scratch on subsets of \ourdataset{} with a fixed timestep $t=0$, varying both the vision features and the dataset size to evaluate their effects on text spotting performance. We use ResNet-50~\cite{he2016deep}, which serves as the vision backbone in the original TESTR~\cite{zhang2022text}, along with diffusion features extracted from Stable Diffusion 2.1\footnote{\texttt{stabilityai/stable-diffusion-2-1-base}} available on Hugging Face. We use model weights trained for 100K steps with a batch size of 8 and a learning rate of 1e-4, using 2 NVIDIA RTX 3090 GPUs. All other implementation details follow the default settings of TESTR~\cite{zhang2022text}.

\subsection{Result Analysis}
As shown in Tab.~\ref{sup:tab:diff_feat}, with a dataset size of 100K, diffusion-based features demonstrate superior recognition performance compared to ResNet~\cite{he2016deep} features. Importantly, when the dataset size is reduced to 20K samples, models relying on ResNet~\cite{he2016deep} features fail to effectively learn recognition, whereas diffusion features still achieve meaningful recognition learning even under limited data conditions. This shows that ResNet~\cite{he2016deep} features require a larger amount of training data to capture text semantics, which is consistent with the reliance on a large-scale synthetic dataset~\cite{liu2020abcnet} in existing text spotting methods~\cite{zhang2022text, ye2023deepsolo, qiao2024dntextspotter, huang2024bridging}. In contrast, diffusion features, pretrained on diverse image-text pairs, can offer improved adaptability to text recognition, even when trained with limited real-world data.

%% file: supple/tab/sup_diff_feat.tex
\begin{table}[h]
    \centering
    \resizebox{1.0\textwidth}{!}{
        \begin{tabular}{ c|l|ccccc } 
                
            \toprule
            \multirow{2}{*}{Dataset Size} & 
            \multirow{2}{*}{Backbone} & 
            \multicolumn{3}{c}{Detection} &
            \multicolumn{2}{c}{End-to-End} \\
            \cmidrule(lr){3-5}
            \cmidrule(lr){6-7}
             &
             &
            Precision($\uparrow$) &
            Recall($\uparrow$)&
            F1-Score($\uparrow$)&
            None($\uparrow$) &
            Full($\uparrow$)
            \\

            \midrule 
            \multirow{2}{*}{$20K$} 

            & ResNet-50~\cite{he2016deep} & 
            {77.67} & {65.91} & 71.31 & {1.81} & {1.96}  \\
            
            & Stable Diffusion~\cite{rombach2022high} & 
            \textbf{82.26} & \textbf{81.78} & \textbf{82.02} & \textbf{27.42} & \textbf{43.87} \\
             
            \midrule 
            \multirow{2}{*}{$100K$}

            & ResNet-50~\cite{he2016deep} & 
            83.24 & 74.52 & 78.64 & 17.03 & {28.70} \\
            
            & Stable Diffusion~\cite{rombach2022high} & 
            \textbf{84.55} & \textbf{82.09} & \textbf{83.30} & \textbf{32.29} & \textbf{45.77} \\
            
        \bottomrule    
        \end{tabular}}

\vspace{5pt}
\caption{\textbf{Comparsion of backbone features for training a text spotting module~\cite{zhang2022text}.} We evaluate the text detection and recognition performance of each trained model on SA-Text test set used in Tab.~\ref{tab:ocr_main_table}.}
\label{sup:tab:diff_feat}
\end{table}

%% file: supple/sec/C_impl_detail.tex
\input{supple/tab/sup_baseline_samtext}
\input{supple/tab/sup_baseline_realtext}

\section{Implementation Details}
\label{sup:C_impl_detail}

%\subsection{Text-Aware Image Restoration (TAIR)}
\paragraph{Evaluation metric.} 
We evaluate text spotting performance~\cite{liu2020abcnet, liu2021abcnet, zhang2022text, huang2024bridging} using standard detection and recognition metrics. For detection, we report Precision (P), Recall (R), and F1-score (F), where a detection is considered correct if its Intersection over Union (IoU) with a ground-truth box exceeds 0.5. For recognition, we adopt two lexicon-based evaluation settings: None and Full. The None setting evaluates recognition without any lexicon, requiring exact matches to ground-truth transcriptions, reflecting performance in open-vocabulary scenarios. The Full setting permits matching predictions to the closest entry in a ground-truth lexicon, simulating closed-vocabulary conditions. This dual evaluation provides a comprehensive assessment of recognition accuracy.

\paragraph{Model overview.}
Given a low-quality (LQ) image $I_{lq} \in \mathbb{R}^{H \times W \times 3}$, the objective is to recover a high-quality (HQ) image $I_{hq} \in \mathbb{R}^{H \times W \times 3}$ with enhanced visual and textual fidelity with $H=W=512$. The LQ image is first processed by a lightweight degradation removal module~\cite{liang2021swinir} and encoded by a VAE~\cite{kingma2013auto} to obtain a conditional latent $c \in \mathbb{R}^{\frac{H}{8} \times \frac{W}{8} \times 4}$. The HQ image is encoded and perturbed with noise to produce a noisy latent $z_t \in \mathbb{R}^{\frac{H}{8} \times \frac{W}{8} \times 4}$. These are channel-wise concatenated to form the input condition $c_t \in \mathbb{R}^{\frac{H}{8} \times \frac{W}{8} \times 8}$. Along with a diffusion timestep $t$ and a text prompt embedding $p_t \in \mathbb{R}^{n \times d}$, where $n = 77$ and $d = 1024$, $c_t$ is fed into the diffusion-based image restoration module.

% Let $\mathrm{DRM}$, $\mathrm{IRM}$, $\mathrm{TSM}$, $\mathrm{VAE}$, and $\mathrm{CLIP}$ denote the light-weight degradation removal module, image restoration module, text spotting module, Variational Autoencoder~\cite{kingma2013auto}, and CLIP text encoder~\cite{radford2021learning}, respectively.
After a single forward pass, we extract four intermediate diffusion features from the four decoder blocks of the U-Net-based restoration module. Each feature is processed by a separate convolutional layer to align the channel dimensions, after which the features are stacked to form a multi-scale diffusion feature input $F \in \mathbb{R}^{L \times D}$ to the transformer-based text spotting module, where $L = 9472$ denotes the total number of stacked tokens and $D = 256$ is the transformer hidden dimension. Deformable attention~\cite{zhu2020deformable} is employed to alleviate the high attention computation cost. 

Following transformer-based text spotting methods~\cite{zhang2022text, qiao2024dntextspotter, huang2024bridging} and inspired by DETR~\cite{carion2020end}, two sets of queries, $q_{\text{det}} \in \mathbb{R}^{Q \times D}$ and $q_{\text{rec}} \in \mathbb{R}^{Q \times D}$ with $Q = 100$, are provided to the detection and recognition decoders $\mathcal{D}^{\text{det}}$ and $\mathcal{D}^{\text{rec}}$, respectively. These queries are processed through cross-attention with the encoder output in the respectful decoder layers. The resulting predictions are denoted as $\{{d^{(i)}}\}_{i=1}^K$ and $\{{r^{(i)}}\}_{i=1}^K$, where $K$ is the number of instances with confidence scores above a threshold $T = 0.5$. Each $d^{(i)} = (d_1^{(i)}, \ldots, d_N^{(i)})$ represents a polygon with $N = 16$ control points, and $r^{(i)} = (r_1^{(i)}, \ldots, r_M^{(i)})$ contains $M = 25$ recognized characters.

\paragraph{Training details.} 
To construct high-quality (HQ) and low-quality (LQ) training pairs for SA-Text, the LQ images were synthesized using the default degradation settings from the Real-ESRGAN pipeline~\cite{wang2018esrgan}. The final model used for performance evaluation in the main paper was trained on four NVIDIA H100 GPUs, with a batch size of 32 per GPU. Each of the three training stages (Stage 1 to Stage 3) was trained for 100,000 iterations. The hyperparameters for the text spotting encoder loss function~\ref{eq:ts_enc}, decoder loss function~\ref{eq:ts_dec}, and stage3 loss function~\ref{eq:stage3} are set as follows: $\lambda_{\text{cls}} = 2.0$, $\lambda_{\text{coord}} = 5.0$, $\lambda_{\text{char}} = 4.0$, $\lambda_{\text{gIoU}} = 2.0$ and $\lambda=0.01$. The number of inference sampling steps for the diffusion based image restoration module was set to $50$.

\textbf{Loss functions.}  
Instance classification employs the focal loss~\cite{lin2017focal}, which is the difference of positive and negative terms to handle class imbalance. For the $j$-th query, the classification loss is given by
$$
\begin{aligned}
\mathcal{L}_{\mathrm{cls}}^{(j)} ={} & - \mathbf{1}_{\{j \in \mathrm{Im}(\sigma)\}} \, \alpha (1 - \hat{b}^{(j)})^{\gamma} \log(\hat{b}^{(j)}) \\
& - \mathbf{1}_{\{j \notin \mathrm{Im}(\sigma)\}} \, (1 - \alpha) (\hat{b}^{(j)})^{\gamma} \log(1 - \hat{b}^{(j)}),
\end{aligned}
$$
where $\hat{b}^{(\cdot)}$ denotes the predicted confidence score, and $\mathrm{Im}(\sigma)$ is the set of indices matched by the optimal bipartite assignment $\sigma$ between predictions and ground truth instances.

The control point regression loss encourages precise localization by minimizing the $\ell_1$ distance between predicted and ground truth control points:
$$
\mathcal{L}_{\mathrm{coord}}^{(j)} = \mathbf{1}_{\{j \in \mathrm{Im}(\sigma)\}} \sum_{i=1}^{N} \left\lVert \hat{d}^{(j)}_i - d^{(\sigma^{-1}(j))}_i \right\rVert_1,
$$
where $\hat{d}_i^{(\cdot)}$ and $d_i^{(\cdot)}$ denote predicted and ground truth control points, respectively, and $\sigma^{-1}$ maps prediction index $j$ to its assigned ground truth index.

Character classification is formulated as a multi-class problem and optimized via cross-entropy loss:
$$
\mathcal{L}_{\mathrm{char}}^{(j)} = \mathbf{1}_{\{j \in \mathrm{Im}(\sigma)\}} \sum_{i=1}^{M} \left( -r^{(\sigma^{-1}(j))}_i \log \hat{r}^{(j)}_i \right),
$$
where $\hat{r}_i^{(\cdot)}$ is the predicted probability for class $i$, and $r_i^{(\cdot)}$ is the one-hot encoded ground truth label.

% Supervision at the encoder employs the same bipartite matching to align bounding box proposals with ground truth. The classification loss $\mathcal{L}_{\mathrm{cls}}^{(i)}$ and control point loss $\mathcal{L}_{\mathrm{coord}}^{(i)}$ correspond to those used for polygon detection, differing only in the matching $\sigma'$. Bounding box regression is guided by the generalized IoU loss $\mathcal{L}_{\mathrm{gIoU}}$ as per~\cite{rezatofighi2019generalized}. The model’s total loss is the sum of encoder and decoder losses.

\paragraph{Text spotting module prompter.}
During inference, given a LQ image, the text spotting module produces $K$ recognized text instances for sampling timestep $t$, denoted as $\{{r^{(i)}}\}_{i=1}^{K}$. These are subsequently passed to the prompter module, $\mathrm{Prompter}(\cdot)$, to generate the input prompt for the image restoration module at the next denoising diffusion timestep $t-1$, formulated as $p_{t-1} = \mathrm{Prompter}(\{{r^{(i)}}\}_{i=1}^{K})$. A handcrafted textual template is employed in the form: “A realistic scene where the texts ‘$\mathrm{text}_1$’, ‘$\mathrm{text}_2$’, … appear clearly on signs, boards, buildings, or other objects.”, where each $\mathrm{text}_i$ corresponds to $r^{(i)}$.

%% file: supple/tab/sup_baseline_samtext.tex
\begin{table}[t]
    \centering
    \resizebox{\textwidth}{!}{
        \begin{tabular}{ c|l|ccccc|ccccc }
            \toprule
            \multirow{3}{*}{\textbf{Deg. Level}} &
            \multirow{3}{*}{\textbf{Model}} &
            \multicolumn{5}{c|}{\textbf{ABCNet v2~\cite{liu2021abcnet}}} &
            \multicolumn{5}{c}{\textbf{TESTR~\cite{zhang2022text}}} \\ 

            \cmidrule(lr){3-7}
            \cmidrule(lr){8-12}
            & & 
            \multicolumn{3}{c}{Detection} &
            \multicolumn{2}{c|}{End-to-End} &
            \multicolumn{3}{c}{Detection} &
            \multicolumn{2}{c}{End-to-End} \\
            
            \cmidrule(lr){3-5}
            \cmidrule(lr){6-7}
            \cmidrule(lr){8-10}
            \cmidrule(lr){11-12}
           
            & & Precision($\uparrow$) & Recall($\uparrow$) & F1-Score($\uparrow$) & None($\uparrow$) & Full($\uparrow$) & Precision($\uparrow$) & Recall($\uparrow$) & F1-Score($\uparrow$) & None($\uparrow$) & Full($\uparrow$) \\ 
            \midrule
            
            % ---------- Level 1 ----------
            \multirow{3}{*}{Level1} 

            & DiffBIR~\cite{lin2024diffbir} & \underline{76.29} & \underline{56.44} & \underline{64.88} & \underline{23.14} & \underline{32.73} & \underline{84.00} & 52.13 & 64.34 & \underline{25.51} & \underline{35.47} \\ 

            & DiffBIR$^\dagger$~\cite{lin2024diffbir} & 53.01 & 51.86 & 52.43 & 15.26 & 20.71 & 60.53 & 51.99 & 55.94& 16.78& 22.82 \\ 
            
            & \textbf{\ourmodel{} (Ours)} & \textbf{85.29} & \textbf{58.34} & \textbf{69.29} & \textbf{26.59} & \textbf{35.69} & \textbf{87.50} & \textbf{54.90} & \textbf{67.47} & \textbf{28.19} & \textbf{36.99} \\ 
            \midrule
            \midrule

            % ---------- Level 2 ----------
            \multirow{3}{*}{Level2} 
            & DiffBIR~\cite{lin2024diffbir} & \underline{72.96} & \underline{53.94} & \underline{62.03} & \underline{19.60} & \underline{27.52} & \underline{79.69} & \underline{50.50} & \underline{61.82} & \underline{21.64} & \underline{30.36} \\ 

            & DiffBIR$^\dagger$~\cite{lin2024diffbir} & 53.28 & 51.18& 52.21& 14.75& 19.61& 58.85& 50.18& 54.17& 16.15& 21.43 \\ 
            
            & \textbf{\ourmodel{} (Ours)} & \textbf{83.02} & \textbf{56.30} & \textbf{67.10} & \textbf{24.42} & \textbf{33.23} & \textbf{86.95} & \textbf{52.86} & \textbf{65.75} & \textbf{26.39} & \textbf{35.13} \\ 
            \midrule
            \midrule

            % ---------- Level 3 ----------
            \multirow{3}{*}{Level3} 

            & DiffBIR~\cite{lin2024diffbir} & \underline{59.30} & \underline{42.20} & \underline{49.31} & \underline{13.88} & \underline{19.39} & \underline{72.27} & \underline{38.98} & \underline{50.65} & \underline{15.61} & \underline{22.67} \\

            & DiffBIR$^\dagger$~\cite{lin2024diffbir} & 44.16& 40.62& 42.31& 10.15& 14.45& 49.73& 41.25& 45.09& 10.75& 15.41 \\ 

            & \textbf{\ourmodel{} (Ours)} & \textbf{81.76} & \textbf{44.11} & \textbf{57.30} & \textbf{19.61} & \textbf{27.50} & \textbf{84.50} & \textbf{42.02} & \textbf{56.13} & \textbf{19.92} & \textbf{28.34} \\ 

            \bottomrule
        \end{tabular}
    }
    \vspace{2pt}
    \caption{\textbf{Text spotting baseline comparison on SA-Text} Each block presents text restoration performance under varying degradation levels, evaluated using two text spotting models~\cite{liu2021abcnet,zhang2022text}. ‘None’ indicates recognition without the use of a lexicon, while ‘Full’ denotes recognition assisted by a full lexicon. The best results are shown in \textbf{bold}, and the second-best are \underline{underlined}.}

    \label{sup:tab_baseline_samtext}
\end{table}

%% file: supple/tab/sup_baseline_realtext.tex
\begin{table}[t]
    \centering
    \resizebox{\linewidth}{!}{
        \begin{tabular}{ l|ccccc|ccccc } 
            \toprule
            \multirow{3}{*}{\textbf{Model}} &
            \multicolumn{5}{c|}{\textbf{ABCNet v2~\cite{liu2021abcnet}}} & 
            \multicolumn{5}{c}{\textbf{TESTR~\cite{zhang2022text}}} \\ 

            \cmidrule(lr){2-6}
            \cmidrule(lr){7-11}
            
            & 
            \multicolumn{3}{c}{Detection} &
            \multicolumn{2}{c|}{End-to-End} &
            \multicolumn{3}{c}{Detection} &
            \multicolumn{2}{c}{End-to-End} \\
            
            \cmidrule(lr){2-4}
            \cmidrule(lr){5-6}
            \cmidrule(lr){7-9}
            \cmidrule(lr){10-11}
        
            & Precision($\uparrow$) & Recall($\uparrow$) & F1-Score($\uparrow$) & None($\uparrow$) & Full($\uparrow$) & Precision($\uparrow$) & Recall($\uparrow$) & F1-score($\uparrow$) & None($\uparrow$) & Full($\uparrow$) \\ 
            \midrule 
            
            DiffBIR~\cite{lin2024diffbir} & \underline{66.04} & \underline{59.69} & \underline{62.71} & \underline{33.75} & \underline{40.05} & \underline{76.33} & \underline{61.87} & \underline{68.35} & \underline{39.27} & \underline{46.11} \\ 

            DiffBIR$^\dagger$~\cite{lin2024diffbir} & 55.31 & 56.02& 55.67& 26.85& 31.23& 58.99& 60.19& 59.58 & 31.41 & 35.98\\

            \textbf{\ourmodel{} (Ours)} & \textbf{83.95} & \textbf{67.58} & \textbf{74.88} & \textbf{48.39} & \textbf{55.01} & \textbf{84.30} & \textbf{67.37} & \textbf{74.89} & \textbf{49.39} & \textbf{56.45} \\ 
            \bottomrule    
        \end{tabular}
    }
    \vspace{3pt}
    \caption{\textbf{Text spotting baseline comparison on Real-Text.} We evaluate the text detection and recognition accuracy of the restored images on Real-Text using text spotting models~\cite{liu2021abcnet,zhang2022text}. None’ indicates recognition without the use of a lexicon, while ‘Full’ denotes
recognition assisted by a full lexicon. The best results are shown in bold, and the second-best are
underlined.}
    \vspace{-5pt}
    \label{sup:tab_baseline_realtext}    
\end{table}

%% file: supple/sec/D_add_quan.tex
\input{supple/tab/sup_abl1}

\section{Additional Quantitative Results}
\label{sup:D_add_quan}

\subsection{Extended Baseline Comparison}
Our main baseline image restoration model is DiffBIR~\cite{lin2024diffbir}. We further include comparisons with $\text{DiffBIR}^{\dagger}$ (denoted as DiffBIR v2.1), which leverages publicly available weights from the official GitHub repository that have undergone additional fine-tuning beyond what was reported in the original publication. The comparative results on text restoration performance are provided in Tab.~\ref{sup:tab_baseline_samtext} for SA-Text and Tab.~\ref{sup:tab_baseline_realtext} for Real-Text.

\subsection{Extended Ablation Experiments}
Extending the analysis in Tab.~\ref{tab:abl} of the main paper, we present additional ablation results in Tab.~\ref{sup:abl}, including evaluations across three degradation levels on SA-Text and on the Real-Text dataset. The table compares (1) two model training stages, referred to as Stage1 and Stage3, and (2) three prompting strategies for the image restoration module: $\mathrm{null}$ (no prompt), $\mathrm{pr}$ (prompter generated prompt), and $\mathrm{gt}$ (ground truth prompt). The ground truth prompt is constructed from the textual content present in the LQ image and follows the format: “A realistic scene where the texts \{$\mathrm{text}_1, \mathrm{text}_2, ...$\} appear clearly on signs, boards, buildings, or other objects.” The prompter-generated prompt adopts the same format, using recognized texts extracted from the LQ image by our text spotting module or a conventionally used LLaVA prompter~\cite{liu2023visual}.

\paragraph{Overall comparison.} 
Comparing $\text{Stage1}_{\mathrm{null}}$ and $\text{Stage3}_{\mathrm{null}}$, we observe accuracy gains of +4.27\%, +2.93\%, +2.43\% and +2.08\% in Tab.~\ref{sup:abl_a} through Tab.~\ref{sup:abl_d}, respectively, indicating the benefit of training with text-aware supervision in Stage3. Next, we compare $\text{Stage1}_{\mathrm{pr}}$ and $\text{Stage3}_{\mathrm{pr}}$, where the image restoration module is guided by prompts generated from the LQ image. In Stage1, we use the LLaVA prompter due to the absence of a text-spotting module, whereas Stage3 leverages our trained text-spotting module prompter. The improved performance of Stage3 indicates that our module provides more accurate and text-aware prompts, enhancing restoration quality. Finally, in the $\text{gt}$ setting, where ground-truth texts are available, both Stage1 and Stage3 show substantial performance improvements with minimal difference between them. This indicates that this ideal scenario in which the restoration module is provided with exact textual information is not achievable in practice, and serves primarily to establish an upper bound on performance.

\paragraph{Comparison on SA-Text degradation levels.}
Comparing $\text{Stage1}_{\mathrm{pr}}$ and $\text{Stage3}_{\mathrm{pr}}$ across the three degradation levels of SA-Text, we observe that the performance gap increases with higher degradation severity. This highlights the importance of Stage3, which learns text-aware features and enables prompting the restoration module using our text-spotting module, proving these prompts' superiority over those from an external LLaVA prompter.

\subsection{User Study Evaluation}
\input{supple/tab/sup_user_study}
To evaluate the quality of both text and image restoration achieved by our \ourmodel{}, we conducted a simple user study comparing it with our baseline, DiffBIR~\cite{lin2024diffbir}. The study included 10 samples: 5 from SA-Text Level 3 and 5 from Real-Text. A total of 21 participants took part in the evaluation. Samples for SA-Text and Real-Text were selected from the examples shown in Tab.~\ref{tab:ocr_main_table} and Tab.~\ref{tab:real_ocr_table}, respectively. As shown in Tab.~\ref{sup:tab:user-study}, the user study results indicate that our \ourmodel{} outperforms the baseline in both text restoration and visual quality. These results highlight that humans often consider text semantics when evaluating image quality, an aspect not fully captured by existing image metrics.

Each participant evaluated the samples based on the following set of questions.:\\
\vspace{-10pt}
\begin{enumerate}
    \item Which image better restores the text content? (Image 1 / Image 2)
    \item Which image better restores the overall appearance? (Image 1 / Image 2)
\end{enumerate}

\input{supple/fig/user_study_example}

%% file: supple/tab/sup_abl1.tex
\begin{table*}[t]
\centering

% First row of tables
\begin{minipage}{0.45\linewidth}
\subfloat[
\textbf{SA-Text (Level 1).} Results evaluated on the SA-Text dataset with Level 1 degradation.
\label{sup:abl_a}
]{
\tablestyle{5pt}{1.0}
\resizebox{\textwidth}{!}{
\begin{tabular}{lccccc}
\toprule
\multirow{3}{*}{\textbf{Model}} & \multicolumn{5}{c}{\textbf{TESTR~\cite{zhang2022text}}} \\
\cmidrule(lr){2-6}
&
\multicolumn{3}{c}{Detection} &
\multicolumn{2}{c}{End-to-End} \\
\cmidrule(lr){2-4}
\cmidrule(lr){5-6}
& Precision($\uparrow$) & Recall($\uparrow$) & F1-Score($\uparrow$) & None($\uparrow$) & Full($\uparrow$) \\
\midrule 
Stage1\textsubscript{$\mathrm{null}$} & 82.09 & 50.27 & 62.36 & 25.36 & 31.66  \\
Stage1\textsubscript{$\mathrm{pr}$} & \underline{83.84} & \underline{52.67} & \underline{64.70} & \underline{27.39} & \underline{35.58}  \\
Stage1\textsubscript{$\mathrm{gt}$} & \textbf{86.79} & \textbf{63.42} & \textbf{73.28} & \textbf{33.68} & \textbf{44.21} \\
\midrule
\midrule
Stage3\textsubscript{$\mathrm{null}$} & 84.68 & \underline{56.12} & \underline{67.50} & 26.94 & 35.93 \\
Stage3\textsubscript{$\mathrm{pr}$} & \textbf{87.50} & 54.90 & 67.47 & \underline{28.19} & \underline{36.99} \\
Stage3\textsubscript{$\mathrm{gt}$} & \underline{86.49} & \textbf{63.28} & \textbf{73.09} & \textbf{34.71} & \textbf{44.87} \\
\bottomrule
\end{tabular}}
}
\end{minipage}
\vspace{2ex}
\hfill
\begin{minipage}{0.45\linewidth}
\subfloat[
\textbf{SA-Text (Level 2).} Results evaluated on the SA-Text dataset with Level 2 degradation.
\label{sup:abl_b}
]{
\tablestyle{5pt}{1.0}
\resizebox{\textwidth}{!}{
\begin{tabular}{lccccc}
\toprule
\multirow{3}{*}{\textbf{Model}} & \multicolumn{5}{c}{\textbf{TESTR~\cite{zhang2022text}}} \\
\cmidrule(lr){2-6}
&
\multicolumn{3}{c}{Detection} &
\multicolumn{2}{c}{End-to-End} \\
\cmidrule(lr){2-4}
\cmidrule(lr){5-6}
\cmidrule(lr){2-4}
\cmidrule(lr){5-6}
& Precision($\uparrow$) & Recall($\uparrow$) & F1-Score($\uparrow$) & None($\uparrow$) & Full($\uparrow$) \\
\midrule
Stage1\textsubscript{$\mathrm{null}$} & 81.77 & 47.37 & 59.99 & 21.24 & 29.79  \\
Stage1\textsubscript{$\mathrm{pr}$} & \underline{82.01} & \underline{49.82} & \underline{61.99} & \underline{24.76} & \underline{31.70}  \\
Stage1\textsubscript{$\mathrm{gt}$} & \textbf{85.09} & \textbf{61.56} & \textbf{71.44} & \textbf{32.51} & \textbf{42.71} \\
\midrule
\midrule
Stage3\textsubscript{$\mathrm{null}$} & 84.47 & \underline{56.21} & \underline{67.50} & 23.46 & 32.72 \\
Stage3\textsubscript{$\mathrm{pr}$} & \textbf{86.95} & 52.86 & 65.75 & \underline{26.39} & \underline{35.13} \\
Stage3\textsubscript{$\mathrm{gt}$} & \underline{86.18} & \textbf{61.60} & \textbf{71.85} & \textbf{33.31} & \textbf{43.40} \\
\bottomrule
\end{tabular}}
}
\end{minipage}

\vspace{2ex}

% Second row of tables
\begin{minipage}{0.45\linewidth}
\subfloat[
\textbf{SA-Text (Level 3).} Results evaluated on the SA-Text dataset with Level 3 degradation.
\label{sup:abl_c}
]{
\tablestyle{5pt}{1.0}
\resizebox{\textwidth}{!}{
\begin{tabular}{lccccc}
\toprule
\multirow{3}{*}{\textbf{Model}} & \multicolumn{5}{c}{\textbf{TESTR~\cite{zhang2022text}}} \\
\cmidrule(lr){2-6}
&
\multicolumn{3}{c}{Detection} &
\multicolumn{2}{c}{End-to-End} \\
\cmidrule(lr){2-4}
\cmidrule(lr){5-6}
\cmidrule(lr){2-4}
\cmidrule(lr){5-6}
& Precision($\uparrow$) & Recall($\uparrow$) & F1-Score($\uparrow$) & None($\uparrow$) & Full($\uparrow$) \\
\midrule
Stage1\textsubscript{$\mathrm{null}$} & 75.69 & 39.53 & 51.94 & 16.32 & 22.22  \\
Stage1\textsubscript{$\mathrm{pr}$} & \underline{75.74} & \underline{41.61} & \underline{53.72} & \underline{18.02} & \underline{24.46}  \\
Stage1\textsubscript{$\mathrm{gt}$} & \textbf{80.60} & \textbf{55.94} & \textbf{66.04} & \textbf{27.08} & \textbf{37.57} \\
\midrule
\midrule
Stage3\textsubscript{$\mathrm{null}$} & \underline{77.04} & \underline{47.91} & \underline{59.08} & 17.44 & 24.65 \\
Stage3\textsubscript{$\mathrm{pr}$} & \textbf{84.50} & 42.02 & 56.13 & \underline{19.92} & \underline{28.34} \\
Stage3\textsubscript{$\mathrm{gt}$} & 80.04 & \textbf{55.44} & \textbf{65.51} & \textbf{27.91} & \textbf{37.76} \\
\bottomrule
\end{tabular}}
}
\end{minipage}
\vspace{1.5ex}
\hfill
\begin{minipage}{0.45\linewidth}
\subfloat[
\textbf{Real-Text.} Results evaluated on the Real-Text dataset.
\label{sup:abl_d}
]{
\tablestyle{5pt}{1.0}
\resizebox{\textwidth}{!}{
\begin{tabular}{lccccc}
\toprule
\multirow{3}{*}{\textbf{Model}} & \multicolumn{5}{c}{\textbf{TESTR~\cite{zhang2022text}}} \\
\cmidrule(lr){2-6}
&
\multicolumn{3}{c}{Detection} &
\multicolumn{2}{c}{End-to-End} \\
\cmidrule(lr){2-4}
\cmidrule(lr){5-6}
\cmidrule(lr){2-4}
\cmidrule(lr){5-6}
& Precision($\uparrow$) & Recall($\uparrow$) & F1-Score($\uparrow$) & None($\uparrow$) & Full($\uparrow$) \\
\midrule
Stage1\textsubscript{$\mathrm{null}$} & 81.91 & 62.68 & 71.02 & 44.11 & 50.70  \\
Stage1\textsubscript{$\mathrm{pr}$} & 81.23 & 65.48 & 72.51 & 44.72 & 51.28  \\
Stage1\textsubscript{$\mathrm{gt}$} & \textbf{83.46} & \textbf{73.61} & \textbf{78.23} & \textbf{52.59} & \textbf{59.28} \\
\midrule
\midrule
Stage3\textsubscript{$\mathrm{null}$} & 81.06 & \underline{70.27} & \underline{75.28} & 46.09 & 52.78 \\
Stage3\textsubscript{$\mathrm{pr}$} & \textbf{84.30} & 67.37 & 74.89 & \underline{49.39} & \underline{56.45} \\
Stage3\textsubscript{$\mathrm{gt}$} & \underline{83.41} & \textbf{75.28} & \textbf{79.14} & \textbf{54.06} & \textbf{60.74} \\
\bottomrule
\end{tabular}}
}
\end{minipage}

\vspace{1.5ex}

\captionsetup{justification=raggedright,singlelinecheck=false}
\caption{\textbf{Additional ablations for SA-Text and Real-Text.} Subscripts $\mathrm{null}$, $\mathrm{pr}$, and $\mathrm{gt}$ indicate the use of a null prompt, a prompt generated by a captioner, and a ground-truth prompt, respectively. Stage1 and Stage3 refer to models trained in Stage 1 and Stage 3, respectively. The $\mathrm{pr}$ prompt is generated by a LLaVA captioner in Stage1 and by our text-spotting module in Stage3.}
\label{sup:abl}
\end{table*}

%% file: supple/tab/sup_user_study.tex
\begin{table}[h]
    \centering
    \resizebox{0.4\textwidth}{!}{
    \begin{tabular}{l|cc}
        \toprule
        \textbf{Criteria} & \textbf{DiffBIR}~\cite{lin2024diffbir} & \textbf{Ours} \\
        \midrule
        Text Quality & 1.50\% & \textbf{98.5\%}   \\
        Image Quality &  11.0\% & \textbf{89.0\%}  \\
        \bottomrule 
    \end{tabular}
    }
    \vspace{5pt}
    \caption{\textbf{User study on text and image restoration quality.}}
    \label{sup:tab:user-study}
    \vspace{-5pt}
\end{table}

%% file: supple/fig/user_study_example.tex
\begin{figure}[h]
    \centering
    \includegraphics[width=0.8\linewidth]{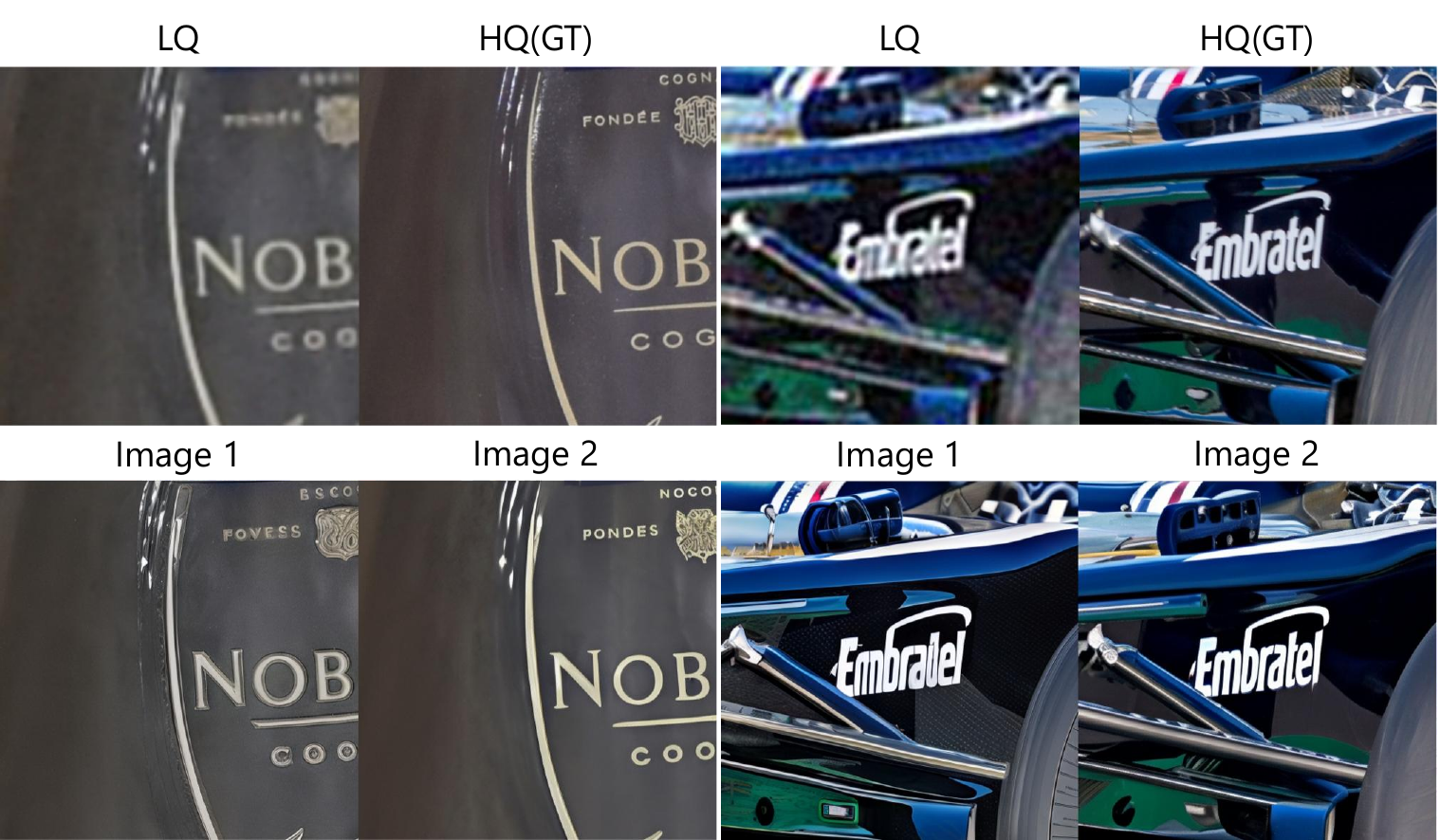}
    \caption{
    \textbf{Example samples for user study.}
    }
    \vspace{-10pt}
    \label{sup:fig_userstudy}
\end{figure}

%% file: supple/sec/E_add_qual.tex
\section{Additional Qualitative Results}
\label{sup:E_add_qual}

In Fig.~\ref{sup:fig_level1}, Fig.~\ref{sup:fig_level2} , Fig.~\ref{sup:fig_level3}, and  Fig.~\ref{sup:fig_realtext} we show further qualitative results on text-aware image restoration (TAIR). The results on \ourdataset{} across different degradation levels and those on Real-Text, demonstrate that our model outperforms other diffusion-based methods in text restoration.

\input{supple/fig/level1_fig}
\input{supple/fig/level2_fig}
\input{supple/fig/level3_fig}
\input{supple/fig/realtext_fig}

%% file: supple/fig/level1_fig.tex
\begin{figure}[h]
    \centering
    \includegraphics[width=\linewidth]{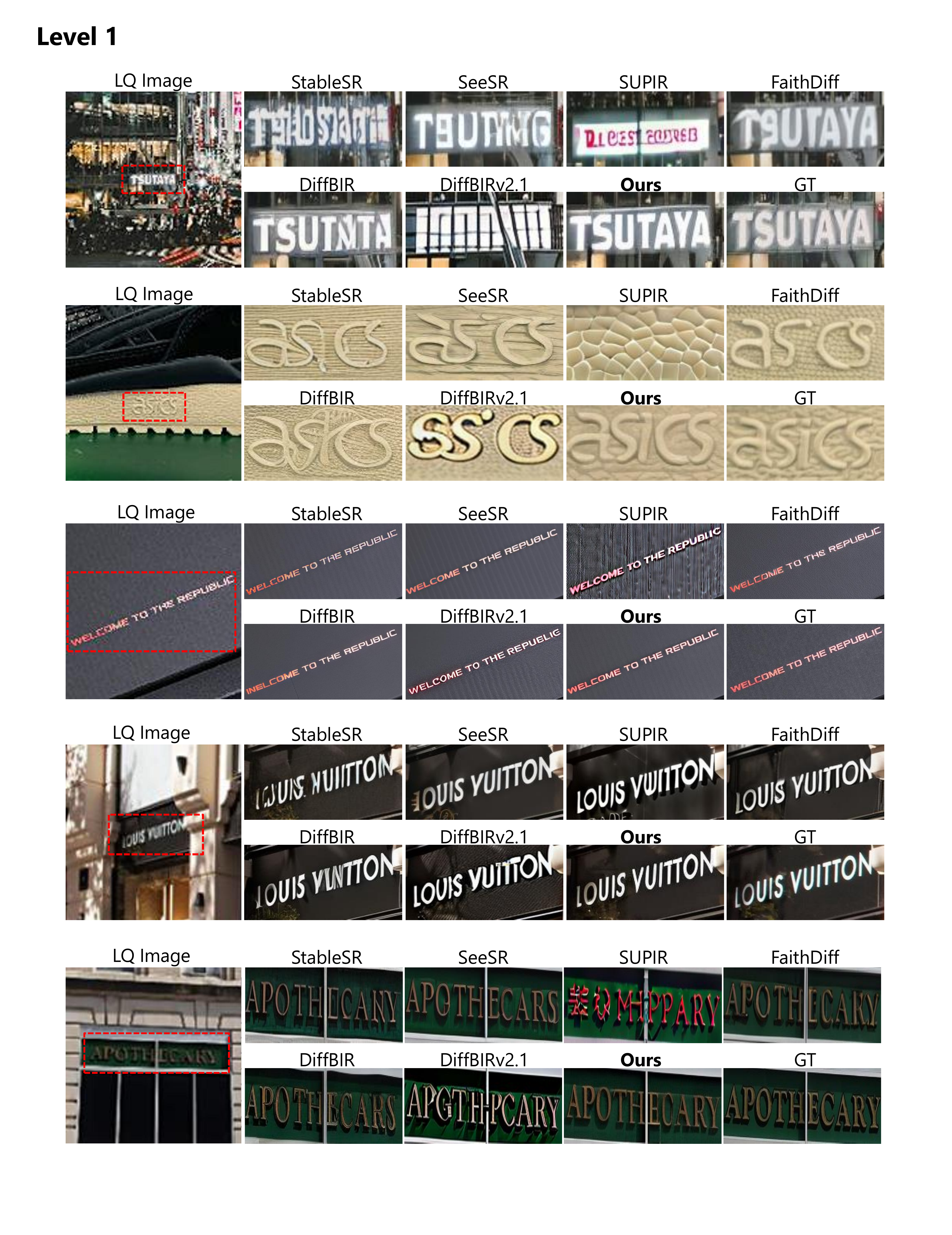}
    \caption{
    \textbf{Qualitative results on \ourdataset{} test set Level 1.}
    }
    \vspace{-10pt}
    \label{sup:fig_level1}
\end{figure}

%% file: supple/fig/level2_fig.tex
\begin{figure}[h]
    \centering
    \includegraphics[width=\linewidth]{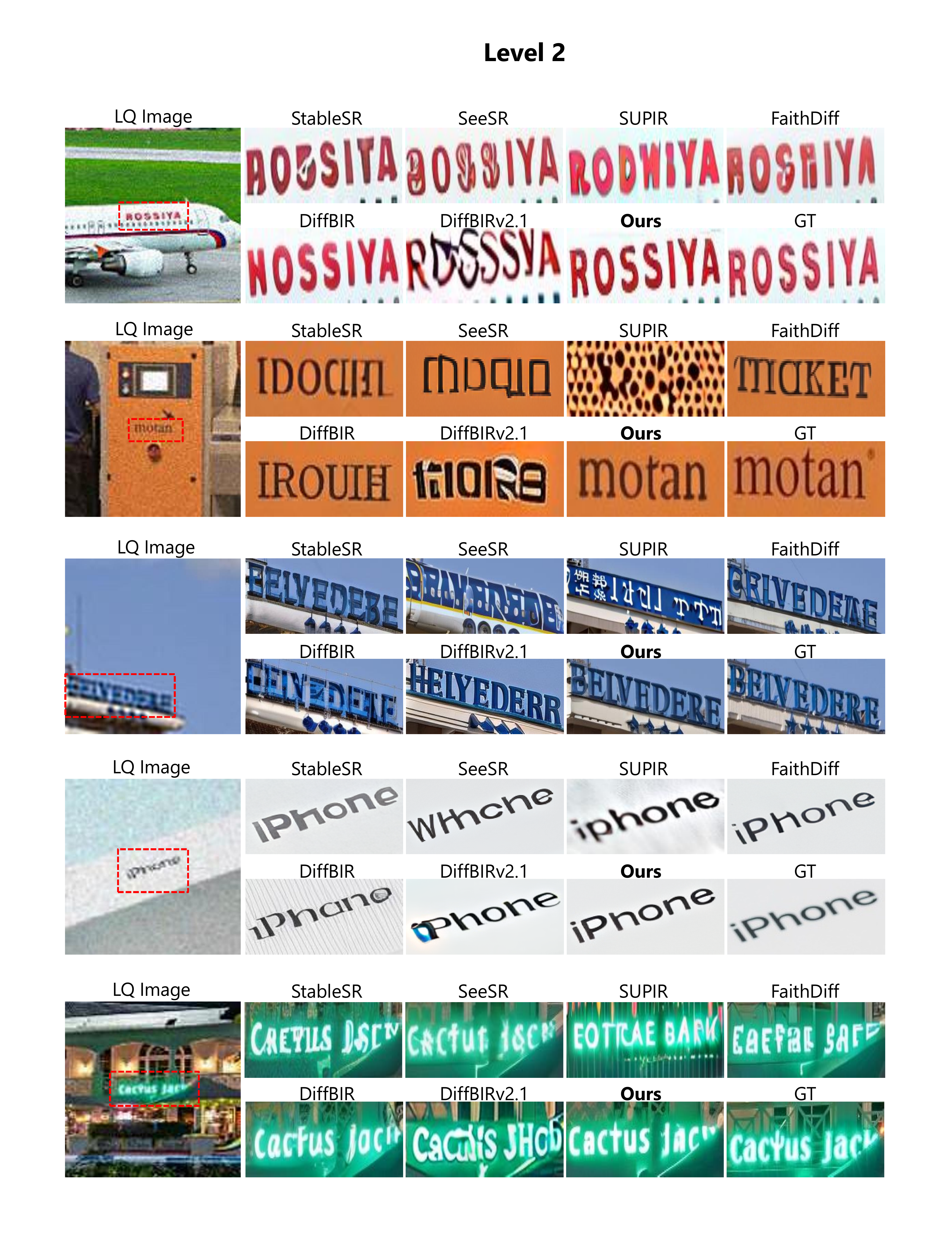}
    \caption{
    \textbf{Qualitative results on \ourdataset{} test set Level 2.}
    }
    \vspace{-10pt}
    \label{sup:fig_level2}
\end{figure}

%% file: supple/fig/level3_fig.tex
\begin{figure}[h]
    \centering
    \includegraphics[width=\linewidth]{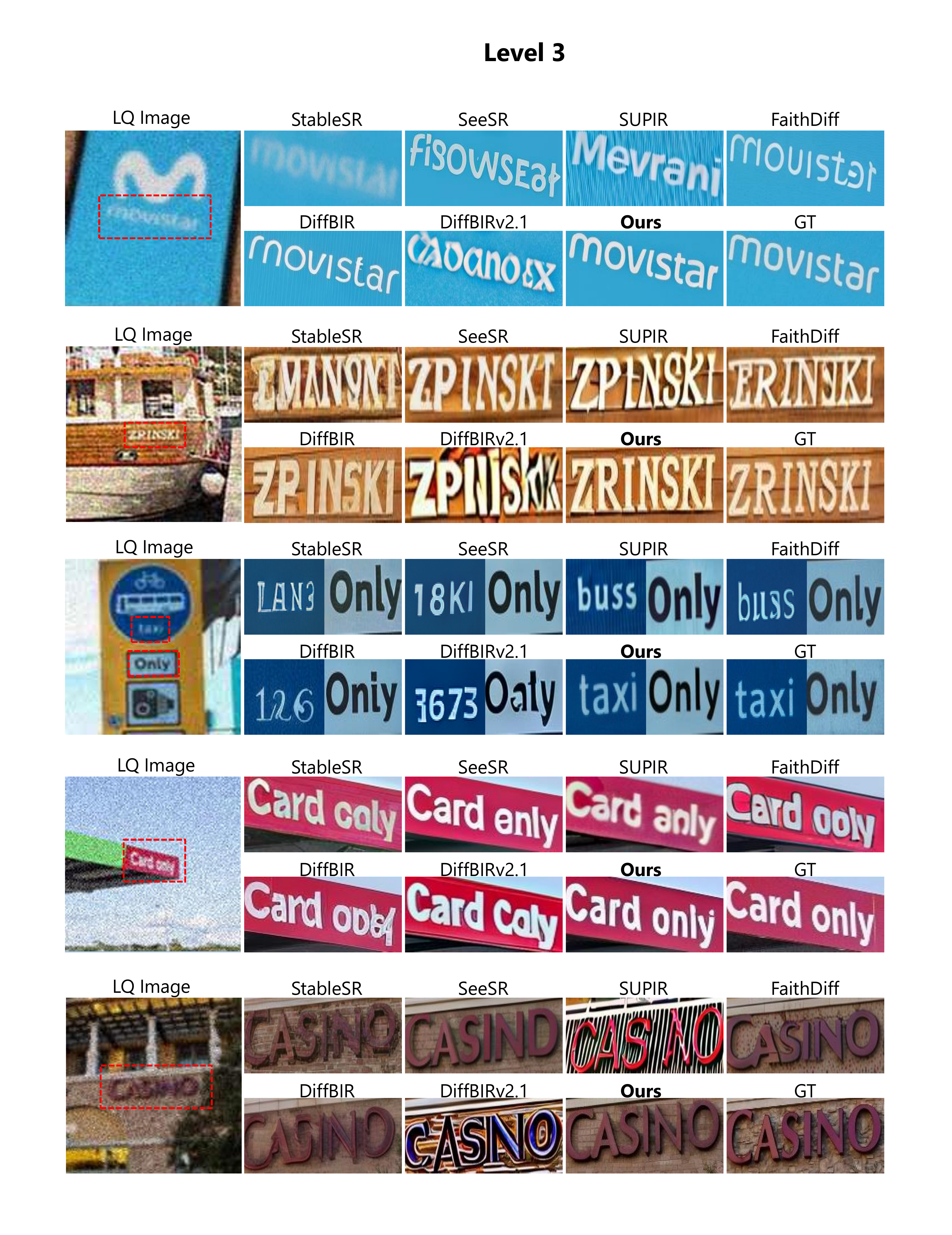}
    \caption{
    \textbf{Qualitative results on \ourdataset{} test set Level 3.}
    }
    \vspace{-10pt}
    \label{sup:fig_level3}
\end{figure}

%% file: supple/fig/realtext_fig.tex
\begin{figure}[h]
    \centering
    \includegraphics[width=\linewidth]{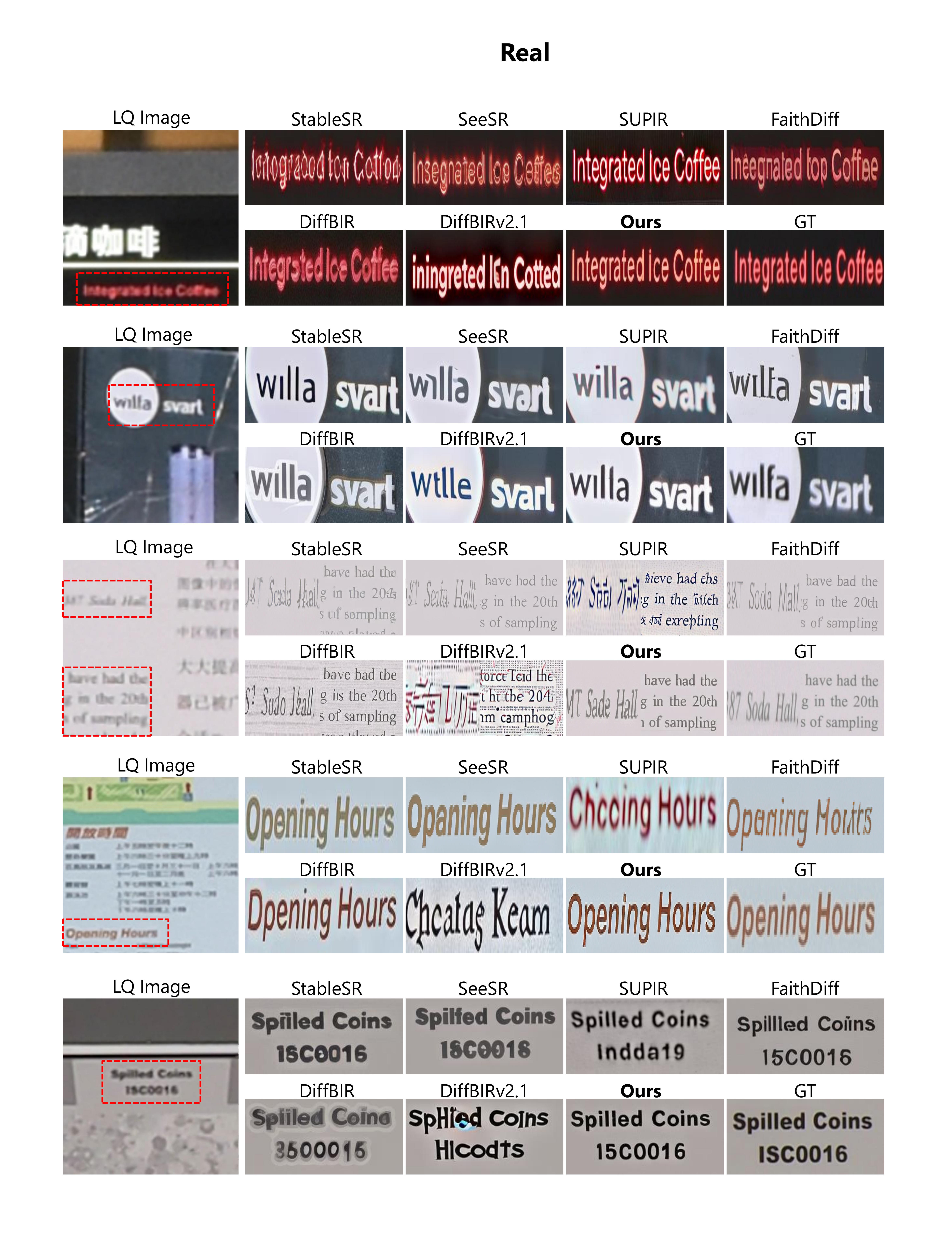}
    \caption{
    \textbf{Qualitative results on Real-Text.}
    }
    \vspace{-10pt}
    \label{sup:fig_realtext}
\end{figure}